\documentclass{article}

\usepackage[english]{babel}

\usepackage[letterpaper,top=2cm,bottom=2cm,left=3cm,right=3cm,marginparwidth=1.75cm]{geometry}

\usepackage{amsmath}
\usepackage{graphicx}

\usepackage{amsfonts}
\usepackage{stfloats}
\usepackage{xspace}
\usepackage{dcolumn}
\usepackage{etoolbox}
\usepackage{caption}
\def\tsc#1{\csdef{#1}{\textsc{\lowercase{#1}}\xspace}}
\tsc{WGM}
\tsc{QE}
\usepackage{amssymb}
\usepackage{amsmath}
\usepackage{threeparttable}
\usepackage{array}
\usepackage{multirow}
\usepackage[table]{xcolor}
\usepackage{xcolor}%
\usepackage{booktabs}
\usepackage{makecell}
\usepackage{longtable,booktabs}
\usepackage{ragged2e}

\usepackage{graphicx}
\usepackage{subcaption}
\usepackage{float}
\usepackage{placeins}
\usepackage{tikz}

\usepackage{algorithm}
\usepackage{algpseudocode}
\usepackage{tcolorbox}
\usepackage[normalem]{ulem}

\newcommand{\SubFigTL}[4]{%
\begin{subfigure}[t]{#1}
    \centering
    \begin{tikzpicture}[baseline=(current bounding box.north)] 
        \node[anchor=south west, inner sep=0] (img) at (0,0)
            {\includegraphics[width=\linewidth]{#2}};
        \node[anchor=north west,
              xshift=6pt, yshift=-4pt,
              font=\bfseries\small,
              fill=white, fill opacity=0.7,
              text opacity=1,
              inner sep=2pt] at (img.north west)
              {#3};
    \end{tikzpicture}
    \label{#4}
\end{subfigure}
}
\usepackage{etoolbox}

\usepackage{colortbl}
\usepackage{pifont}
\definecolor{stdFuchsia}{RGB}{255, 0, 255}
\definecolor{stdSilver}{RGB}{192, 192, 192}
\definecolor{lightBlue}{RGB}{224, 255, 255}
\definecolor{gold}{RGB}{255, 223, 0}

\definecolor{myred}{RGB}{255, 182, 193}
\definecolor{myblue}{RGB}{173, 216, 230}

\newenvironment{mytable}[1][htbp]
  {\begin{table}[#1]\centering\small}
  {\end{table}}

\usepackage{todonotes,tikz}
\usetikzlibrary{arrows.meta, shapes.geometric, positioning}
\usetikzlibrary{decorations.pathreplacing, calligraphy}
\usetikzlibrary{shapes, fit, snakes, quotes, angles, calc}

\usepackage[colorlinks=true, allcolors=blue]{hyperref}

\newcommand{\CNS}{\text{CNS}}
\newcommand{\SOFA}{\text{SOFA}}

\begin{document}

\title{CNA: An AI-Oriented Comprehensive Normalized Assessment for Healthy Status and Application to Optimize RRT Strategies by Reinforcement Learning\footnote{This is the author’s accepted manuscript of an article published in Expert Systems with Applications 332 (2027) 133464. The final published version is available at \url{https://doi.org/10.1016/j.eswa.2026.133464}.}}

\author{Jiang Liu$^{1,2,\ast}$, Chan Zhou$^{1,2,\ast}$, Yujie Li$^{3}$, Di Wu$^{4}$, Yihao Xie$^{1,2}$, Peiwei Li$^{1,2}$, \\
Xin Shu$^{3}$, Jiaqi Zhu$^{5}$, Chunyong Yang$^{3}$, Yuwen Chen$^{1,2}$, and Bin Yi$^{3,\dagger}$}

\date{}

\maketitle

\begin{center}
\footnotesize
\begin{tabular}{@{}l@{}}
$^{1}$Chongqing Institute of Green and Intelligent Technology, CAS \\
$^{2}$Chongqing School, University of Chinese Academy of Sciences \\
$^{3}$Department of Anesthesiology, Southwest Hospital, Third Military Medical University (Army Medical University) \\
$^{4}$College of Computer and Information Science, Southwest University\\
$^{5}$Institute of Software, Chinese Academy of Sciences \\
$^{\ast}$ Co-first authors of this manuscript. Both of them contributed equally to this work. \\
$^{\dagger}$ Corresponding author.
\end{tabular}
\end{center}

\begin{abstract}
Millions worldwide require Renal Replacement Therapy (RRT) as a treatment essential for survival. However, optimizing RRT strategies via AI is challenging due to heterogeneous patient dynamics, missing data, and the absence of an AI-oriented health assessment criterion. We propose an AI-Oriented Comprehensive Normalized Assessment (CNA) for healthy status and apply it to optimize RRT strategies by using offline reinforcement learning (RL). The key idea of CNA is transforming vital-sign distributions into a standard normal space, enabling a unified, data-driven health-status score defined by deviations from referent intervals, which also provides an AI-oriented criterion to assess strategy quality and supports RL termination. We further design a structured 23-dimensional state representation that integrates 19 indicators with 4 RRT descriptors, and employ matrix decomposition to reconstruct missing vital signs, improving data completeness for learning. These components are incorporated into multiple offline RL algorithms and validated via systematic ablation studies on RRT feature subsets. Compared with physicians’ observed treatments, the best learned strategy reduces mortality from 13.2\% to 5.0\% (reducing 62.24\%) and shortens average in-hospital stay from 308.5 to 250.1 hours (reducing 18.93\%), demonstrating both methodological innovation and the potential of CNA-guided RL to improve RRT outcomes in nephrology.
\end{abstract}


\section{Introduction}
\label{sec1}
Artificial Intelligence is demonstrating potential in nephrology, with applications spanning kidney segmentation, functional assessment, and disease diagnosis \cite{xu2024narrative}. In the domain of Renal Replacement Therapy (RRT), specifically Continuous RRT (CRRT), AI is being explored for predicting clinical outcomes such as Acute Kidney Injury (AKI) requiring RRT \cite{kelly2021potential}, prognosticating mortality and kidney recovery \cite{hammouda2022can}, and optimizing treatment delivery \cite{yoo2023predicting}. However, many of these innovations remain in early development stages, necessitating further validation and robust implementation studies \cite{leung2024deep}.

The selection of RRT modality: CRRT, Intermittent RRT (IRRT), or no treatment, is a critical decision impacting patient survival, renal recovery, healthcare costs, and quality of life \cite{schoenfelder2017effects},  \cite{lins2009intermittent}. RL is a subfield of machine learning particularly suited for such long-term, sequential decision-making problems under uncertainty \cite{wang2024clinical}, \cite{qayyum2020secure}. Unlike supervised or unsupervised learning, RL learns optimal strategies through environmental interaction and evaluative feedback without requiring a pre-defined model of the environment \cite{lee2018machine}. This capability to optimize for long-term rewards makes it highly applicable to dynamic clinical settings like RRT management \cite{ando2005framework}, \cite{khamis2014adaptive}, \cite{lewis2012reinforcement}. 

Therefore, this study employs RL to optimize RRT strategies. A critical aspect of RL is designing an appropriate reward function \cite{kuhnle2021designing}. In principle, the reward function should reflect the status change of the patient. While using established medical scores related to RRT is a straightforward approach, like in \cite{liu2024value}, multiple such scores exist in critical care and nephrology, each with strengths and weaknesses depending on the RRT phase. These include the Sequential Organ Failure Assessment (SOFA \cite{vincent1996sofa}, \cite{lambden2019sofa}), Acute Physiology and Chronic Health Evaluation II (APACHE II \cite{knaus1985apache}, \cite{hou2023effects}), KDIGO classification for AKI \cite{awdishu2025kdigo}, \cite{kotani2019modification}, Simplified Acute Physiology Score II (SAPS II \cite{le1993new}, \cite{bae2008continuous}), and the older RIFLE \cite{ricci2008rifle} and AKIN \cite{mehta2007acute} criteria. To address the limitations of individual scores, this article proposes a new comprehensive normalized assessment by considering the factors of those scores. Another significant challenge in RL is determining training termination or convergence, particularly for models like Deep Q-Network (DQN) \cite{mnih2013playing} and Double DQN (DDQN) \cite{hasselt2010double} that operate in continuous state spaces, where convergence is not easily discernible \cite{kuang2018performance}. The termination criterion is equivalent to accurately and reasonably evaluating the performance of AI strategies developed through RL. However, such evaluation remains a great challenge \cite{lins2009intermittent}. To address the issue, we adopt the data-driven AI strategy assessment method TECM*, which is based on the action similarity in \cite{liu2024value}. Given that real-world clinical data often contains missing values, we utilize matrix factorization \cite{nguyen2019low}, \cite{wu2019deep}, \cite{song2019tensor} for missing value reconstruction to ensure robust analysis. 

%
\begin{figure}[H]
	\centering
	\includegraphics[width=0.98\linewidth]{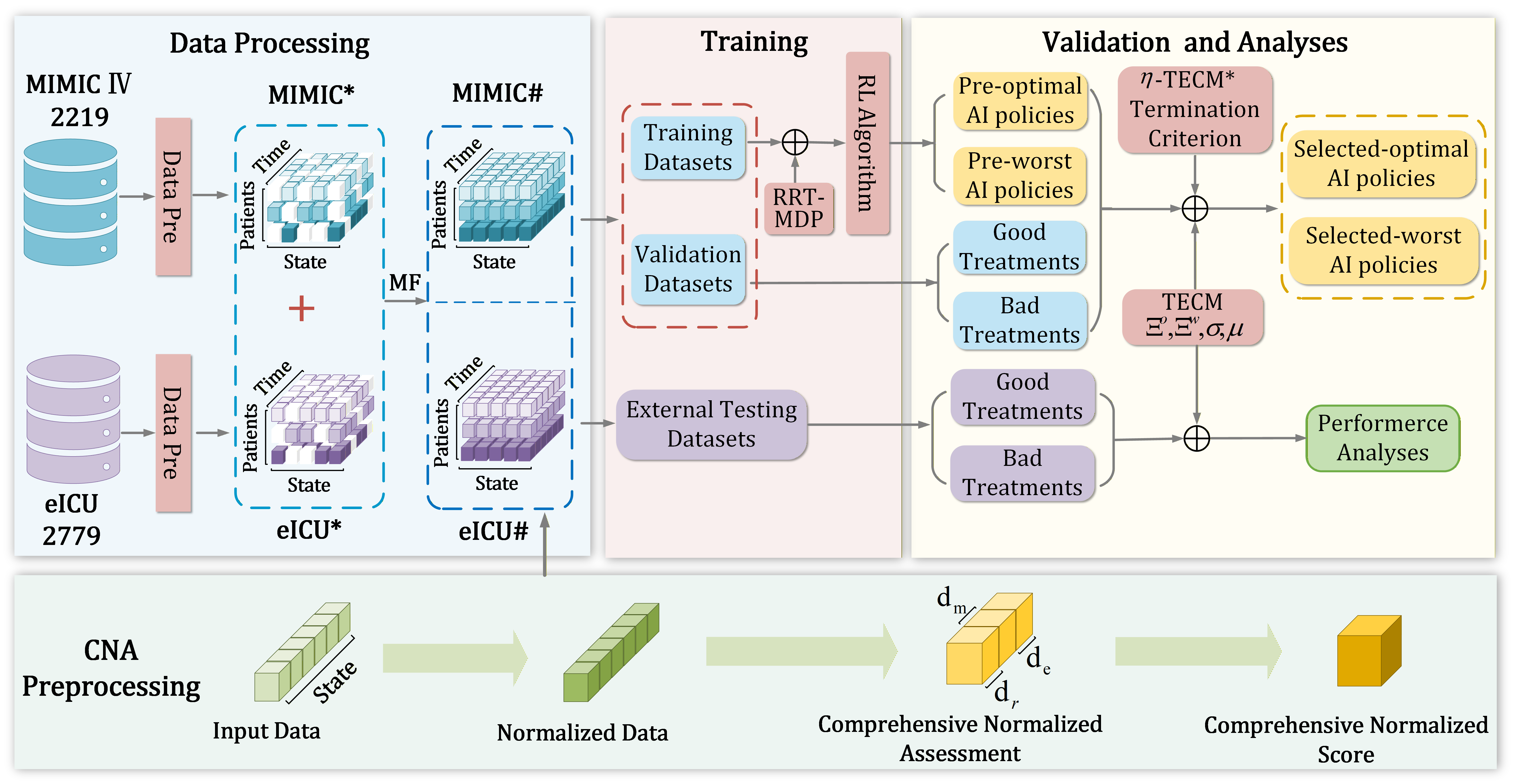}
	\caption{Overview of the new framework. It comprises four main modules: 1) data processing, 2) reinforcement learning (RL) training, 3) model selection based on the Treatment Effect Comparison Matrix (TECM), and 4) results analysis. ``Data Pre" denotes operations performed on initial data including cleansing, preliminary imputation, and window segmentation; MF represents matrix factorization, for data reconstruction. CNA is the new notion of Comprehensive Normalized Assessment proposed in this study.  TECM and $\eta$-TECM* are performance metrics of RL algorithms posed in \cite{liu2025tecm}.}
	\label{fig:framework}
\end{figure}   

In the present study, we propose an RL-based framework as Fig.\ref{fig:framework} for the optimization of RRT treatment strategy. To the end, we construct an MDP model  RRT-MDP for the RRT treatment process. For the real-world clinical data, matrix factorization is used as a data preprocessing module to reconstruct missing values. The primary methodological contribution of this work is the Comprehensive Normalized Assessment (CNA) and its scalar form, the Comprehensive Normalized Score (CNS), which provide a fine-grained health-status assessment and a reward-design basis for reinforcement learning. Four reinforcement learning algorithms, including DQN, DDQN, BCQ \cite{BEOLET2024102920}, \cite{fujimoto2019off}, and CQL \cite{kumar2020conservative}, are then employed to evaluate the applicability of the proposed framework. TECM* \cite{liu2025tecm} is used as a strategy evaluation and training-termination criterion rather than as the central methodological contribution of this study.  
Accordingly, the main contributions of this study are summarized as follows:
\begin{enumerate}
    \item Developing an effective MDP model for RRT treatment.
    \item Using matrix factorization to reconstruct missing data values.
    \item Proposing a Comprehensive Normalized Assessment (CNA) for healthcare AI.
\end{enumerate}

\section{Materials and methods}
\label{sec2}
\subsection{Data preparation}
\label{subsec1}

\begin{figure}[H]
    \centering
    \includegraphics[width=0.8\linewidth]{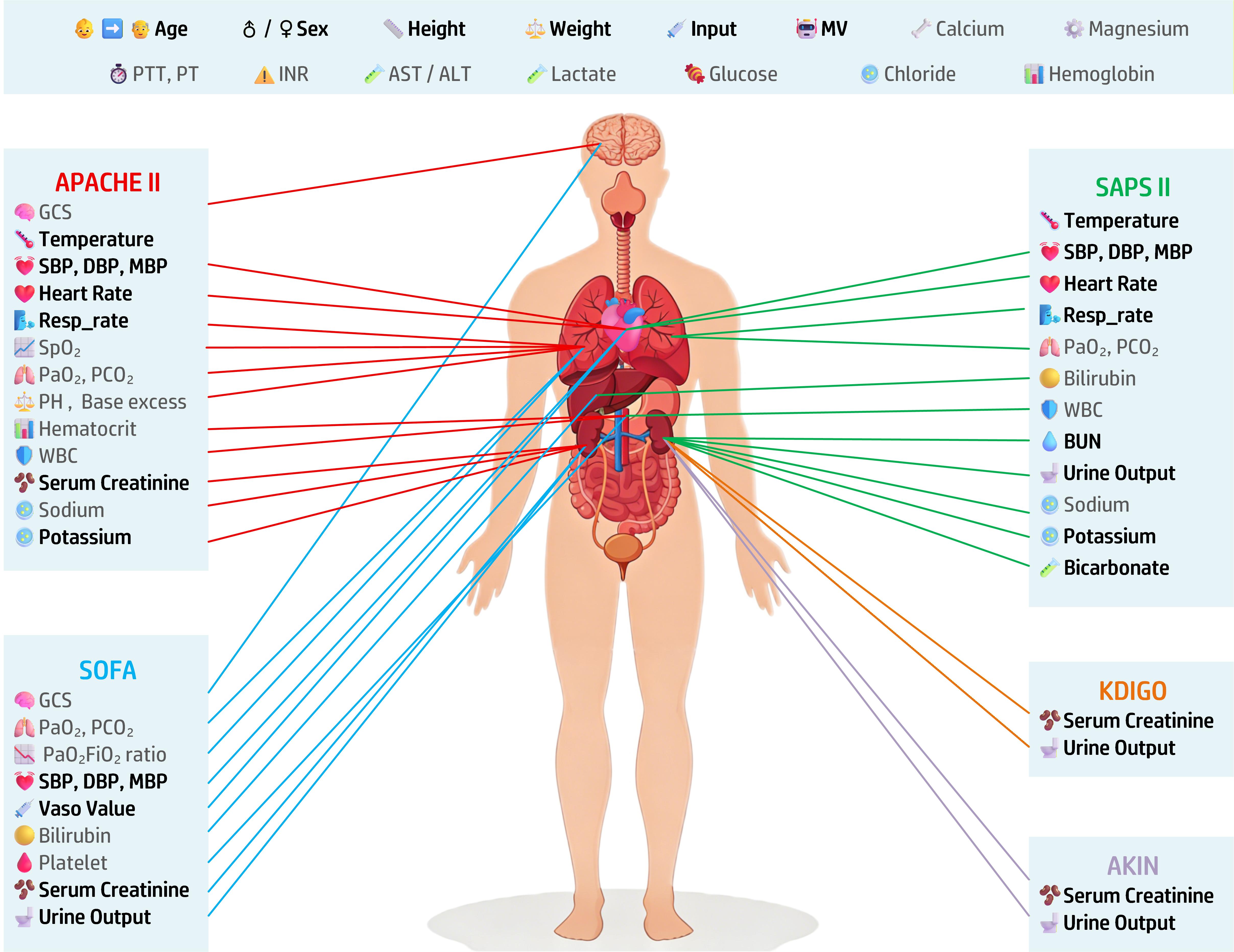}
    \caption{Indicators selected in various medical scores and guidelines }
    \label{fig:state_vital}
\end{figure}

First, it determines which vital signs should be selected for RRT. There is no widely accepted consensus about the selection of vital signs: SAPS II \cite{le1993new} is used in \cite{bae2008continuous}, AKIN \cite{BEOLET2024102920} is used in \cite{ostermann2016acute}, SOFA \cite{vincent1996sofa} is used in \cite{lewis2012reinforcement}, KDIGO AKI \cite{schoenfelder2017effects} is used in \cite{kotani2019modification}, \cite{le1993new}, APACHE II \cite{knaus1985apache} is used in \cite{hou2023effects}. To comprehensively capture the features used in the literature, we collect the vital signs from KDIGO guidelines and pertinent research. The selected indicators are listed in  Table \ref{tab:missing_rates} and Fig. \ref{fig:state_vital}.

\begin{longtable}{@{}lccccccc@{}}
    \caption{ Data information}\label{tab:missing_rates}\\
    \toprule
        \multirow{2}{*}{\centering Indicators} & \multirow{2}{*}{\centering Ref-Vals} & \multicolumn{2}{c}{MIMIC-IV} & \multicolumn{2}{c}{eICU} &  \multirow{2}{*}{Rate} \\
    \cmidrule(lr){3-4} \cmidrule(lr){5-6}
         & & $miss_1$ & $rate_1$ & $miss_2$ & $rate_2$ & \\
    \midrule
    \endfirsthead
    \multicolumn{7}{c}{{\footnotesize \tablename\ \thetable{} -- Continued from previous page}} \\
    \toprule
     \multirow{2}{*}{\centering Indicators} & \multirow{2}{*}{\centering Ref-Vals} & \multicolumn{2}{c}{MIMIC-IV} & \multicolumn{2}{c}{eICU} &  \multirow{2}{*}{Rate} \\
    \cmidrule(lr){3-4} \cmidrule(lr){5-6}
      & & $miss_1$ & $rate_1$ & $miss_2$ & $rate_2$ & \\
    \midrule
    \endhead
    \bottomrule
    \multicolumn{7}{r}{{\footnotesize Continued on next page}} \\
    \endfoot
    \bottomrule
    \multicolumn{7}{@{}p{\dimexpr\textwidth-4\tabcolsep}@{}}{
    \parbox{0.94\columnwidth}{
\vspace{4pt}
The items are listed according to the Rate values. The symbol $*$ means the value unavailable. Ref-Vals denote the reference values. A total of 160,087 and 131,771 records were selected from MIMIC-IV and eICU, respectively. $\text{miss}_i$ denotes the number of missing values for the associated vital sign,  $\text{rate}_1 = \frac{\text{miss}_1}{160,087}$ and $\text{rate}_2 = \frac{\text{miss}_2}{131,771}$ denote the missing rates. Rate $= \frac{\text{miss}_1 + \text{miss}_2}{160,087 + 131,771}$.
   } } \\
\endlastfoot
\rowcolor{gold!10}
Age & $*$ & 0 & 0.00\% & 0 & 0.00\% & 0.00\% \\
\rowcolor{gold!10}
Sex & $*$ & 0 & 0.00\% & 0 & 0.00\% & 0.00\% \\
\rowcolor{gold!10}
MV & $*$ & 0 & 0.00\% & 0 & 0.00\% & 0.00\% \\
\rowcolor{gold!10}
Vaso\_Value & $*$ & 0 & 0.00\% & 0 & 0.00\% & 0.00\% \\
\rowcolor{gold!10}
Weight & $*$ & 0 & 0.00\% & 0 & 0.00\% & 0.00\% \\
\rowcolor{gold!10}
Height & $*$ & 0 & 0.00\% & 0 & 0.00\% & 0.00\% \\
\rowcolor{gold!10}
Input & $*$ & 29792 & 18.61\% & 48085 & 36.49\% & 26.68\% \\
\rowcolor{gold!10}
Resp\_rate & $[12,20]$ & 48542 & 30.32\% & 30588 & 23.21\% & 27.11\% \\
\rowcolor{gold!10}
Heartrate & $[60,100]$ & 48502 & 30.30\% & 33308 & 25.28\% & 28.03\% \\
\rowcolor{gold!10}
Temperature & $[36,37]$ & 53471 & 33.40\% & 32072 & 24.34\% & 29.30\% \\
\rowcolor{gold!10}
Spo2 & $[95,100]$ & 48813 & 30.49\% & 36803 & 27.93\% & 29.33\% \\
\rowcolor{gold!10}
Dbp & $[60,90]$ & 48681 & 30.41\% & 49269 & 37.39\% & 33.56\% \\
\rowcolor{gold!10}
Sbp & $[90,140]$ & 48677 & 30.41\% & 49285 & 37.40\% & 33.56\% \\
\rowcolor{gold!10}
Mbp & $[70,105]$ & 48634 & 30.38\% & 57747 & 43.82\% & 36.45\% \\
\rowcolor{gold!10}
GCS & 15 & 48876 & 30.53\% & 64448 & 48.91\% & 38.83\% \\
\rowcolor{lightBlue!50}
Glucose & $[7,110]$ & 105308 & 65.78\% & 43632 & 33.11\% & 51.03\% \\
\rowcolor{lightBlue!50}
Potassium & $[3.5,5]$ & 104021 & 64.98\% & 51297 & 38.93\% & 53.22\% \\
\rowcolor{lightBlue!50}
Sodium & $[135,145]$ & 103988 & 64.96\% & 51738 & 39.26\% & 53.35\% \\
\rowcolor{lightBlue!50}
Chloride & $[98,106]$ & 104100 & 65.03\% & 51967 & 39.44\% & 53.47\% \\
\rowcolor{lightBlue!50}
Bicarbonate & $[22,27]$ & 104544 & 65.30\% & 52227 & 39.63\% & 53.71\% \\
\rowcolor{lightBlue!50}
Bun & $[9,20]$ & 105052 & 65.62\% & 51963 & 39.43\% & 53.79\% \\
\rowcolor{lightBlue!50}
Creatinine & $[0.5,1.5]$ & 105163 & 65.69\% & 52011 & 39.47\% & 53.85\% \\
\rowcolor{lightBlue!50}
Calcium & $[9,11]$ & 106923 & 66.79\% & 52237 & 39.64\% & 54.53\% \\
\rowcolor{lightBlue!50}
Hematocrit & $[37,50]$ & 112860 & 70.50\% & 53885 & 40.89\% & 57.13\% \\
\rowcolor{lightBlue!50}
Hemoglobin & $[11,16]$ & 115200 & 71.96\% & 53765 & 40.80\% & 57.89\% \\
\rowcolor{lightBlue!50}
Platelet & $[100,300]$ & 115067 & 71.88\% & 54555 & 41.40\% & 58.12\% \\
\rowcolor{lightBlue!50}
Wbc & $[4,10]$ & 115683 & 72.26\% & 54674 & 41.49\% & 58.36\% \\
\rowcolor{lightBlue!50}
UO24 & $[1,3]$ & 94119 & 58.79\% & 94867 & 71.99\% & 64.75\% \\
\rowcolor{lightBlue!50}
Magnesium & $[1.7,2.4]$ & 105638 & 65.99\% & 114026 & 86.53\% & 75.26\% \\
\rowcolor{lightBlue!50}
Inr & $[0.8,1.3]$ & 123198 & 76.96\% & 98750 & 74.94\% & 76.05\% \\
\rowcolor{lightBlue!50}
Pco2 & $[35,45]$ & 107810 & 67.34\% & 118885 & 90.22\% & 77.67\% \\
\rowcolor{lightBlue!50}
Po2 & $[60,100]$ & 107810 & 67.34\% & 118883 & 90.22\% & 77.67\% \\
\rowcolor{lightBlue!50}
PH & $[7.35,7.45]$ & 107812 & 67.35\% & 118906 & 90.24\% & 77.68\% \\
\rowcolor{lightBlue!50}
Baseexcess & $[-3,3]$ & 107810 & 67.34\% & 119679 & 90.82\% & 77.95\% \\
\rowcolor{lightBlue!50}
PTT & $[23,27]$ & 122171 & 76.32\% & 109912 & 83.41\% & 79.52\% \\
\rowcolor{lightBlue!50}
AST & $[9,40]$ & 141758 & 88.55\% & 92906 & 70.51\% & 80.40\% \\
\rowcolor{lightBlue!50}
Bilirubin & $[0.3,1.3]$ & 141705 & 88.52\% & 93316 & 70.82\% & 80.53\% \\
\rowcolor{lightBlue!50}
ALT & $[5,40]$ & 142005 & 88.70\% & 93731 & 71.13\% & 80.77\% \\
\rowcolor{lightBlue!50}
Pao2fio2ratio & $>$300 & 115915 & 72.41\% & 120654 & 91.50\% & 81.03\% \\
\rowcolor{lightBlue!50}
Lactate & $[9,20]$ & 125124 & 78.16\% & 117238 & 88.97\% & 83.04\% \\
\rowcolor{lightBlue!50}
PT & $[135,145]$ & 123198 & 76.96\% & 122877 & 93.25\% & 84.31\% \\
\midrule
\multicolumn{2}{c}{Summary} & 3536428 & 51.37\% & 2644543 & 46.67\% & 49.25\% \\
\end{longtable}

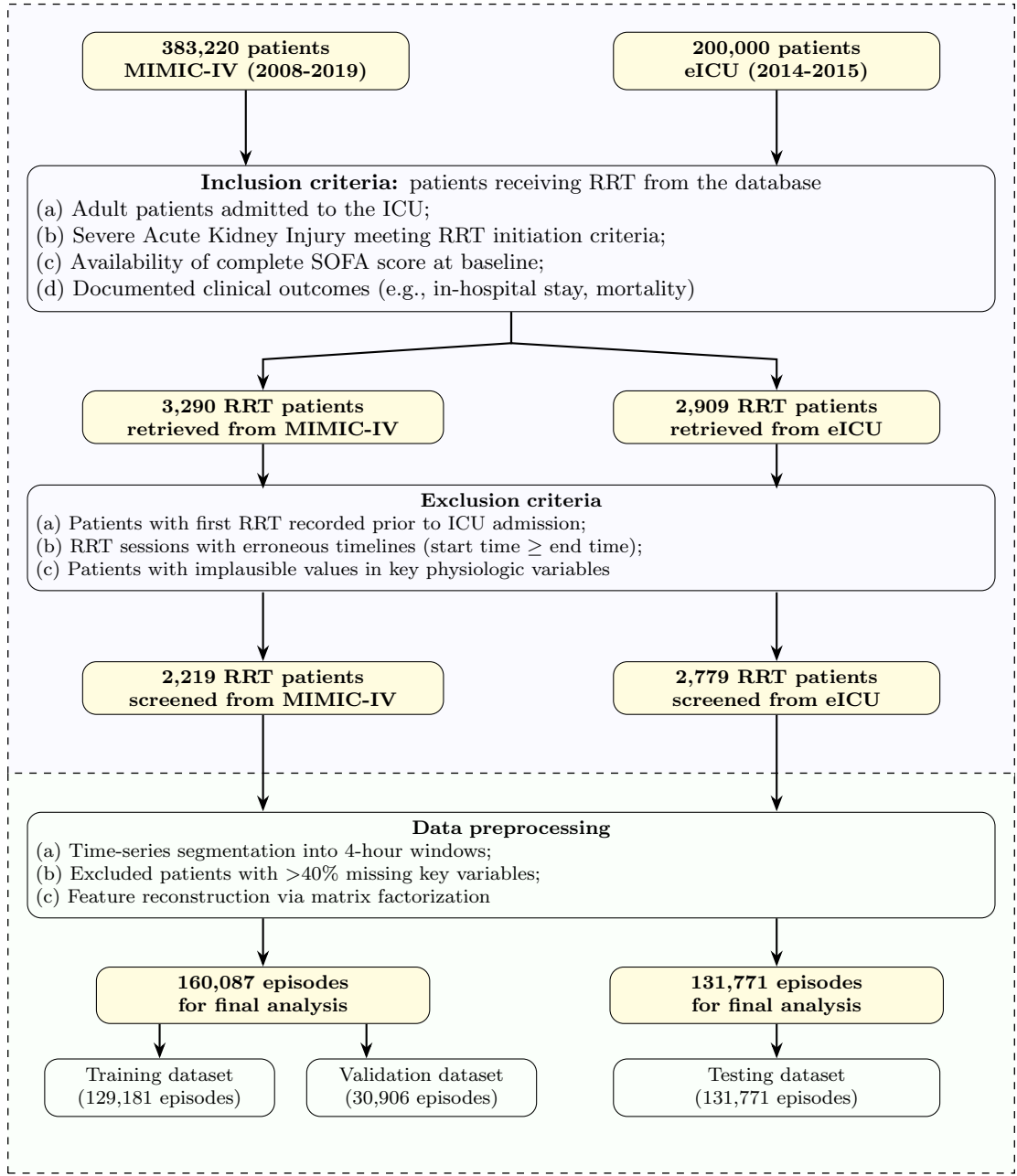
\begin{figure}[!htbp]
\centering
\resizebox{0.95\textwidth}{!}{
\begin{tikzpicture}[
  node distance=0.8cm and 3.5cm,
  box/.style={rectangle, draw, rounded corners=2mm, text width=3cm, 
              text centered, minimum height=0.7cm, font=\footnotesize, align=center},
  header/.style={rectangle, draw,  rounded corners=2mm, 
                text width=3cm, text centered, minimum height=0.8cm, 
                font=\small, align=center},
  data/.style={rectangle, draw, fill=yellow!15, rounded corners=2mm, 
              text width=3cm, text centered, minimum height=0.7cm, 
              font=\footnotesize\bfseries, align=center},
  arrow/.style={->, thick, >=Stealth}
]

\node (abovebox) [draw, align=center, dashed, line width=0.5pt,  rectangle, fill=blue!2,   minimum width = 14.6cm, minimum height = 11.2cm,  shape border rotate=90, aspect=0.08,  xshift=0cm, yshift=-2.2cm ] { };

\node (belowbox) [draw, align=center, dashed, line width=0.5pt,  rectangle,  fill=green!2, minimum width = 14.6cm, minimum height = 5.8cm,  shape border rotate=90, aspect=0.08,  xshift=0cm, yshift=-10.65cm] { };

\node[header, align=left,  text width=13.8cm] (title) at (0, 0) 
  {\centering {\bfseries Inclusion criteria:} patients receiving RRT from the database \\ 
  \begin{tabular}[t]{@{}l@{}}  
  (a)  Adult patients admitted to the ICU; \\
  (b) Severe Acute Kidney Injury meeting RRT initiation criteria; \\
   (c) Availability of complete SOFA score at baseline;\\
   (d) Documented clinical outcomes (e.g., in-hospital stay, mortality)
   \end{tabular}};

\node[box, align=left,  below=of title,  xshift=-0cm , yshift=-1.7cm , text width=13.8cm ] (excl) 
{\centering {\bfseries Exclusion criteria} \\
  \begin{tabular}[t]{@{}l@{}} 
  (a)  Patients with first RRT recorded prior to ICU admission; \\
  (b) RRT sessions with erroneous timelines
(start time $\ge$ end time); \\
  (c) Patients with implausible values in key
physiologic variables 
  \end{tabular}};

\node[box, align=left,  below=of excl, xshift=-0cm , yshift=-2.4cm, text width=13.8cm ] (exc2) 
 {\centering {\bfseries  Data preprocessing }\\
  \begin{tabular}[t]{@{}l@{}} 
  (a) Time-series segmentation into 4-hour windows; \\
  (b) Excluded patients with $>$40\% missing key variables; \\
  (c) Feature reconstruction via matrix factorization
  \end{tabular}};
  
\node[data, above = 3.0cm of title.west, anchor=north west, xshift=0.8 cm, text width= 4.5 cm] (MIMIC) 
  {383,220 patients \\MIMIC-IV (2008-2019)};

\node[data, below=2.2cm of title.west, anchor=north west, xshift=0.8 cm, text width= 5 cm] (source1) 
  {3,290 RRT patients \\ retrieved from MIMIC-IV };

\node[data, below=of source1 , yshift=-2.35cm, text width= 5 cm  ] (analysis1) 
  {2,219 RRT patients \\ screened from MIMIC-IV };

\node[data, below=of analysis1,  yshift=-2.85 cm, text width=4.6 cm  ] (episode1) 
  { 160,087 episodes\\ for final analysis};

\node[box, below=0.5cm of episode1, xshift= -1.5cm,  text width= 3.1 cm ] (train) 
  {Training dataset \\(129,181 episodes)};

\node[box, below=0.5cm of  episode1, xshift= 2.3 cm,  text width= 3.1 cm ] (valid) 
  {Validation dataset \\(30,906 episodes)};

\node[data, above = 3.0cm of title.east, anchor=north east, xshift=-0.8cm,  yshift=-0cm, text width= 4.5 cm] (eICU) 
  {200,000 patients \\
  eICU (2014-2015)};

\node[data, below=2.2cm of title.east, anchor=north east, xshift=-0.8cm,  yshift=-0cm, text width= 4.5 cm] (source2) 
  {2,909 RRT patients \\ retrieved from eICU };

\node[data, below=of source2,  yshift=-2.35cm, text width=4.5 cm  ] (analysis2) 
  {2,779 RRT patients \\ screened from eICU};

\node[data, below=of analysis2,  yshift=-2.85 cm, text width=4.6 cm ] (episode2) 
  {{\bfseries 131,771 episodes } \\for final analysis};

\node[box, below=0.5cm of episode2, xshift = 0.0 cm, text width= 4.5 cm ] (test) 
  { Testing dataset \\(131,771 episodes)};

\draw[arrow] (MIMIC.south) -- ([yshift=-1.1cm]MIMIC.south);
\draw[arrow] (title.south)  -- ([yshift=-0.45cm]title.south) -- ([yshift=0.45cm]source1.north) -- (source1.north) ;
\draw[arrow] (source1.south) -- ([yshift=-0.65cm]source1.south);
\draw[arrow] ([yshift=1.0cm]analysis1.north) -- (analysis1.north);
\draw[arrow] (analysis1.south) -- ([yshift=-1.4cm]analysis1.south);
\draw[arrow] ([yshift=0.7cm]episode1.north) -- (episode1.north);

\draw[arrow] ([yshift =0.5cm ]train.north) -- (train.north);
\draw[arrow] ([xshift =-0.8cm, yshift =0.5cm ]valid.north) -- ([xshift =-0.8cm]valid.north);

\draw[arrow] (eICU.south) -- ([yshift=-1.1cm]eICU.south);
\draw[arrow] (title.south)  -- ([yshift=-0.45cm]title.south) -- ([yshift=0.45cm]source2.north) -- (source2.north) ;
\draw[arrow] (source2.south) -- ([yshift=-0.65cm]source2.south);
\draw[arrow] ([yshift=1.0cm]analysis2.north) -- (analysis2.north);
\draw[arrow] (analysis2.south) -- ([yshift=-1.4cm]analysis2.south);
\draw[arrow] ([yshift=0.7cm]episode2.north) -- (episode2.north);
\draw[arrow] ([yshift=0.5cm]test.north) -- (test.north);
\end{tikzpicture}
}
\caption{Flow diagram of the patient enrollment and data preprocessing}
\label{fig:export}
\end{figure}

This study is based on two publicly available real-world clinical databases, MIMIC-IV \cite{johnson2023mimic} and eICU \cite{li2020optimizing}. Although these datasets are publicly accessible, they are derived from real clinical records rather than simulated, synthetic, or AI-generated data. We used these public databases to enhance the transparency, reproducibility, and comparability of the proposed framework, allowing other researchers to examine the data processing pipeline and benchmark the results under comparable settings. The patient selection process is detailed in Fig. \ref{fig:export}. This study screens 2,219 and 2,779 RRT patients from MIMIC-IV  and eICU, respectively.

Considering the delayed onset of heparin’s therapeutic effect \cite{raghu2017deep}, we discretized patient trajectories using a 4-hour time window  where the data of a patient within each time window were treated as one record. The 4-hour interval was selected for both data-driven \cite{peine2021development}, \cite{kapral2025optimal} and clinical reasons \cite{zhang2025long}. From the data perspective, RRT-related records in the databases were available or could be aggregated at clinically meaningful intervals such as 4, 8, 12, 16, and 24 hours; among these options, 4 hours provided the finest usable temporal resolution while maintaining sufficient data density for trajectory construction. From the clinical perspective, consultation with clinicians from multiple dialysis centers suggested that 4 hours is a reasonable minimum interval for reassessing RRT-related interventions in routine practice. Therefore, the 4-hour window was adopted as a balance between temporal resolution, data completeness, and clinical workflow. Patient data were subsequently represented as multidimensional discrete time series at this temporal resolution, resulting in 287,266 records from 2,219 RRT patients in MIMIC-IV and 214,687 records from 2,779 RRT patients in eICU. 

After segmenting the time windows, we performed data cleaning by removing records with a missing value ratio exceeding 40\%, resulting in 160,087 and 131,771 valid episodes, respectively. Note that each episode consists of timely successive records of a patient, and thus the records of a patient might be broken into several episodes. The selected RRT records from MIMIC-IV and eICU were combined only for the matrix-factorization-based missing-value reconstruction stage.  Specifically, the two datasets formed an $m \times n$ matrix $M$ with missing entries, where $m = 160, 087 + 131, 771 = 291, 858$ and $n=41$, where each row of $M$ corresponds to the record of a patient at a time $t$, which indicates the patient's health status $s_t$ at time $t$, and each column of $M$ corresponds to the values of a fixed indicator for all patients at all times.

After missing-value reconstruction, MIMIC-IV and eICU were separated again for model development and evaluation. The MIMIC-IV cohort was divided at the patient level into training and validation sets with an 8:1 ratio, ensuring that records from the same patient did not appear in both sets. The eICU cohort was reserved as an independent testing dataset. Therefore, model training, validation, hyperparameter selection, and strategy selection were conducted only on the MIMIC-IV training and validation data, while the eICU data were used only for testing after model development.

\subsection{Data imputation method}
\label{subsec2}
Following data cleaning, we used matrix factorization to reconstruct missing values in selected clinical variables. The rationale for adopting matrix factorization is that the selected physiological indicators are not independent, but are jointly generated from the same underlying patient condition. Therefore, latent correlations among vital signs, laboratory measurements, and treatment-related variables can be exploited to estimate partially missing entries in a unified data-driven framework. Matrix factorization has also been widely used for missing-value reconstruction in incomplete biomedical and multivariate data \cite{saha2017effective}, \cite{wu2015post}.

To determine which variables should be reconstructed, two criteria were established for matrix factorization based on feature missing rates in Table \ref{tab:missing_rates}: 1) features with global missing rates $<40\%$ across both MIMIC and eICU datasets; 2) features clinically critical for RRT optimization per Standard Operating Procedures with missing rates \(\leq 70\%\). The original matrix \(M\) was subsequently decomposed into: \(M^{c}\) containing selected features for imputation, and \(M^{d}\) comprising static features (e.g., age, weight) requiring no processing and high-missing-rate indicators. This process can be formally defined as: $M = \mathrm{cp}\left( \left[ M^{c}, M^{d} \right] \right)$, where \(\mathrm{cp}(\cdot)\) denotes a column permutation on the input.

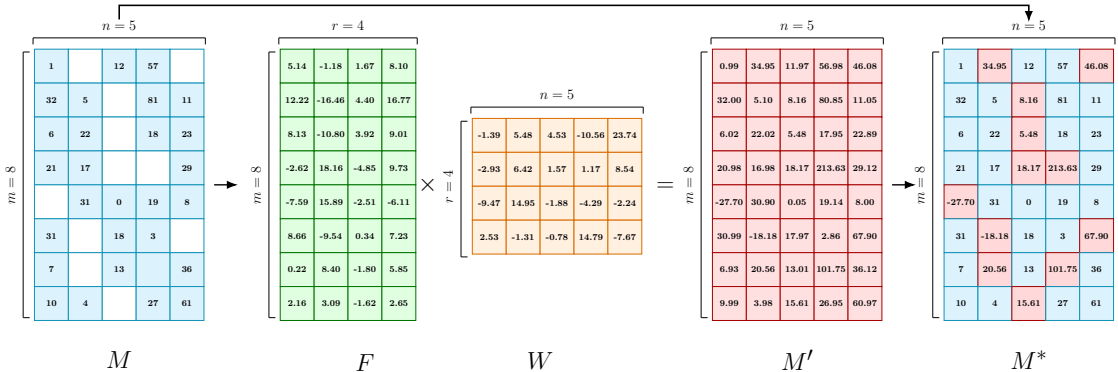
\begin{figure}[!hp]
\centering
\resizebox{0.95\textwidth}{!}{
\begin{tikzpicture}[
    cellblue/.style={
        draw=cyan!70!black,
        fill=cyan!12,
        minimum width=\cw cm,
        minimum height=\ch cm,
        font=\small\bfseries,
        inner sep=0pt
    },
    cellblueempty/.style={
        draw=cyan!70!black,
        fill=white,
        minimum width=\cw cm,
        minimum height=\ch cm,
        font=\small\bfseries,
        inner sep=0pt
    },
    cellgreen/.style={
        draw=green!45!black,
        fill=green!12,
        minimum width=\cw cm,
        minimum height=\ch cm,
        font=\small\bfseries,
        inner sep=0pt
    },
    cellyellow/.style={
        draw=orange!85!black,
        fill=orange!12,
        minimum width=\cw cm,
        minimum height=\ch cm,
        font=\small\bfseries,
        inner sep=0pt
    },
    cellred/.style={
        draw=red!65!black,
        fill=red!12,
        minimum width=\cw cm,
        minimum height=\ch cm,
        font=\small\bfseries,
        inner sep=0pt
    },
    cellfilled/.style={
        draw=red!65!black,
        fill=red!15,
        minimum width=\cw cm,
        minimum height=\ch cm,
        font=\small\bfseries,
        inner sep=0pt
    },
    matrixlabel/.style={
        font=\Huge\bfseries
    },
    dimlabel/.style={
        font=\Large\bfseries
    },
    arrow/.style={
        -{Latex[length=4mm,width=3mm]},
        line width=1.4pt
    }
]

\def\cw{1.2}
\def\ch{1.2}

\coordinate (M) at (0,0);

\node[cellblue]      at ($(M)+(0*\cw,0*\ch)$) {1};
\node[cellblueempty] at ($(M)+(1*\cw,0*\ch)$) {};
\node[cellblue]      at ($(M)+(2*\cw,0*\ch)$) {12};
\node[cellblue]      at ($(M)+(3*\cw,0*\ch)$) {57};
\node[cellblueempty] at ($(M)+(4*\cw,0*\ch)$) {};

\node[cellblue]      at ($(M)+(0*\cw,-1*\ch)$) {32};
\node[cellblue]      at ($(M)+(1*\cw,-1*\ch)$) {5};
\node[cellblueempty] at ($(M)+(2*\cw,-1*\ch)$) {};
\node[cellblue]      at ($(M)+(3*\cw,-1*\ch)$) {81};
\node[cellblue]      at ($(M)+(4*\cw,-1*\ch)$) {11};

\node[cellblue]      at ($(M)+(0*\cw,-2*\ch)$) {6};
\node[cellblue]      at ($(M)+(1*\cw,-2*\ch)$) {22};
\node[cellblueempty] at ($(M)+(2*\cw,-2*\ch)$) {};
\node[cellblue]      at ($(M)+(3*\cw,-2*\ch)$) {18};
\node[cellblue]      at ($(M)+(4*\cw,-2*\ch)$) {23};

\node[cellblue]      at ($(M)+(0*\cw,-3*\ch)$) {21};
\node[cellblue]      at ($(M)+(1*\cw,-3*\ch)$) {17};
\node[cellblueempty] at ($(M)+(2*\cw,-3*\ch)$) {};
\node[cellblueempty] at ($(M)+(3*\cw,-3*\ch)$) {};
\node[cellblue]      at ($(M)+(4*\cw,-3*\ch)$) {29};

\node[cellblueempty] at ($(M)+(0*\cw,-4*\ch)$) {};
\node[cellblue]      at ($(M)+(1*\cw,-4*\ch)$) {31};
\node[cellblue]      at ($(M)+(2*\cw,-4*\ch)$) {0};
\node[cellblue]      at ($(M)+(3*\cw,-4*\ch)$) {19};
\node[cellblue]      at ($(M)+(4*\cw,-4*\ch)$) {8};

\node[cellblue]      at ($(M)+(0*\cw,-5*\ch)$) {31};
\node[cellblueempty] at ($(M)+(1*\cw,-5*\ch)$) {};
\node[cellblue]      at ($(M)+(2*\cw,-5*\ch)$) {18};
\node[cellblue]      at ($(M)+(3*\cw,-5*\ch)$) {3};
\node[cellblueempty] at ($(M)+(4*\cw,-5*\ch)$) {};

\node[cellblue]      at ($(M)+(0*\cw,-6*\ch)$) {7};
\node[cellblueempty] at ($(M)+(1*\cw,-6*\ch)$) {};
\node[cellblue]      at ($(M)+(2*\cw,-6*\ch)$) {13};
\node[cellblue]      at ($(M)+(3*\cw,-6*\ch)$) {};
\node[cellblue]      at ($(M)+(4*\cw,-6*\ch)$) {36};

\node[cellblue]      at ($(M)+(0*\cw,-7*\ch)$) {10};
\node[cellblue]      at ($(M)+(1*\cw,-7*\ch)$) {4};
\node[cellblueempty] at ($(M)+(2*\cw,-7*\ch)$) {};
\node[cellblue]      at ($(M)+(3*\cw,-7*\ch)$) {27};
\node[cellblue]      at ($(M)+(4*\cw,-7*\ch)$) {61};
\node[matrixlabel] at ($(M)+(2*\cw,-8.65*\ch)$) {$M$};

\draw[line width=1pt] ($(M)+(-0.65,1.0)$) -- ($(M)+(4*\cw+0.65,1.0)$);
\draw[line width=1pt] ($(M)+(-0.65,1.0)$) -- ($(M)+(-0.65,0.77)$);
\draw[line width=1pt] ($(M)+(4*\cw+0.65,1.0)$) -- ($(M)+(4*\cw+0.65,0.77)$);

\node[dimlabel] at ($(M)+(2*\cw,1.45)$) {$n=5$};

\draw[line width=1pt] ($(M)+(-1,0.5)$) -- ($(M)+(-1,-7*\ch-0.5)$);
\draw[line width=1pt] ($(M)+(-1,0.5)$) -- ($(M)+(-0.77,0.5)$);
\draw[line width=1pt] ($(M)+(-1,-7*\ch-0.5)$) -- ($(M)+(-0.77,-7*\ch-0.5)$);
\node[dimlabel, rotate=90] at ($(M)+(-1.45,-3.5*\ch)$) {$m=8$};

\draw[arrow] ($(M)+(4.8*\cw,-3.5*\ch)$) -- ++(0.85,0);

\coordinate (F) at (8.7,0);

\foreach \i/\a/\b/\c/\d in {
    0/5.14/-1.18/1.67/8.10,
    1/12.22/-16.46/4.40/16.77,
    2/8.13/-10.80/3.92/9.01,
    3/-2.62/18.16/-4.85/9.73,
    4/-7.59/15.89/-2.51/-6.11,
    5/8.66/-9.54/0.34/7.23,
    6/0.22/8.40/-1.80/5.85,
    7/2.16/3.09/-1.62/2.65
}{
    \node[cellgreen] at ($(F)+(0*\cw,-\i*\ch)$) {\a};
    \node[cellgreen] at ($(F)+(1*\cw,-\i*\ch)$) {\b};
    \node[cellgreen] at ($(F)+(2*\cw,-\i*\ch)$) {\c};
    \node[cellgreen] at ($(F)+(3*\cw,-\i*\ch)$) {\d};
}

\node[matrixlabel] at ($(F)+(2*\cw,-8.75*\ch)$) {$F$};

\draw[line width=1pt] ($(F)+(-0.65,1.0)$) -- ($(F)+(3*\cw+0.65,1.0)$);
\draw[line width=1pt] ($(F)+(-0.65,1.0)$) -- ($(F)+(-0.65,0.77)$);
\draw[line width=1pt] ($(F)+(3*\cw+0.65,1.0)$) -- ($(F)+(3*\cw+0.65,0.77)$);
\node[dimlabel] at ($(F)+(1.5*\cw,1.45)$) {$r=4$};

\draw[line width=1pt] ($(F)+(-1,0.5)$) -- ($(F)+(-1,-7*\ch-0.5)$);
\draw[line width=1pt] ($(F)+(-1,0.5)$) -- ($(F)+(-0.77,0.5)$);
\draw[line width=1pt] ($(F)+(-1,-7*\ch-0.5)$) -- ($(F)+(-0.77,-7*\ch-0.5)$);
\node[dimlabel, rotate=90] at ($(F)+(-1.45,-3.5*\ch)$) {$m=8$};

\node[font=\Huge\bfseries] at ($(F)+(3.85*\cw,-3.5*\ch)$) {$\times$};

\coordinate (W) at (15.5,-2.45);

\foreach \i/\a/\b/\c/\d/\e in {
    0/-1.39/5.48/4.53/-10.56/23.74,
    1/-2.93/6.42/1.57/1.17/8.54,
    2/-9.47/14.95/-1.88/-4.29/-2.24,
    3/2.53/-1.31/-0.78/14.79/-7.67
}{
    \node[cellyellow] at ($(W)+(0*\cw,-\i*\ch)$) {\a};
    \node[cellyellow] at ($(W)+(1*\cw,-\i*\ch)$) {\b};
    \node[cellyellow] at ($(W)+(2*\cw,-\i*\ch)$) {\c};
    \node[cellyellow] at ($(W)+(3*\cw,-\i*\ch)$) {\d};
    \node[cellyellow] at ($(W)+(4*\cw,-\i*\ch)$) {\e};
}

\node[matrixlabel] at ($(W)+(1.5*\cw,-6.65*\ch)$) {$W$};

\draw[line width=1pt] ($(W)+(-0.65,1.0)$) -- ($(W)+(4*\cw+0.65,1)$);
\draw[line width=1pt] ($(W)+(-0.65,1.0)$) -- ($(W)+(-0.65,0.77)$);
\draw[line width=1pt] ($(W)+(4*\cw+0.65,1)$) -- ($(W)+(4*\cw+0.65,0.77)$);
\node[dimlabel] at ($(W)+(2*\cw,1.45)$) {$n=5$};

\draw[line width=1pt] ($(W)+(-1,0.5)$) -- ($(W)+(-1,-4*\ch+0.5)$);
\draw[line width=1pt] ($(W)+(-1,0.5)$) -- ($(W)+(-0.77,0.5)$);
\draw[line width=1pt] ($(W)+(-1,-4*\ch+0.5)$) -- ($(W)+(-0.77,-4*\ch+0.5)$);
\node[dimlabel, rotate=90] at ($(W)+(-1.45,-1.5*\ch)$) {$r=4$};

\node[font=\Huge\bfseries] at ($(W)+(5.2*\cw,-1.5*\ch)$) {$=$};

\coordinate (Mp) at (24.0,0);

\foreach \i/\a/\b/\c/\d/\e in {
    0/0.99/34.95/11.97/56.98/46.08,
    1/32.00/5.10/8.16/80.85/11.05,
    2/6.02/22.02/5.48/17.95/22.89,
    3/20.98/16.98/18.17/213.63/29.12,
    4/-27.70/30.90/0.05/19.14/8.00,
    5/30.99/-18.18/17.97/2.86/67.90,
    6/6.93/20.56/13.01/101.75/36.12,
    7/9.99/3.98/15.61/26.95/60.97
}{
    \node[cellred] at ($(Mp)+(0*\cw,-\i*\ch)$) {\a};
    \node[cellred] at ($(Mp)+(1*\cw,-\i*\ch)$) {\b};
    \node[cellred] at ($(Mp)+(2*\cw,-\i*\ch)$) {\c};
    \node[cellred] at ($(Mp)+(3*\cw,-\i*\ch)$) {\d};
    \node[cellred] at ($(Mp)+(4*\cw,-\i*\ch)$) {\e};
}

\node[matrixlabel] at ($(Mp)+(2*\cw,-8.65*\ch)$) {$M'$};

\draw[line width=1pt] ($(Mp)+(-0.65,1.0)$) -- ($(Mp)+(4*\cw+0.65,1.0)$);
\draw[line width=1pt] ($(Mp)+(-0.65,1.0)$) -- ($(Mp)+(-0.65,0.77)$);
\draw[line width=1pt] ($(Mp)+(4*\cw+0.65,1.0)$) -- ($(Mp)+(4*\cw+0.65,0.77)$);
\node[dimlabel] at ($(Mp)+(2*\cw,1.45)$) {$n=5$};

\draw[line width=1pt] ($(Mp)+(-1,0.5)$) -- ($(Mp)+(-1,-7*\ch-0.5)$);
\draw[line width=1pt] ($(Mp)+(-1,0.5)$) -- ($(Mp)+(-0.77,0.5)$);
\draw[line width=1pt] ($(Mp)+(-1,-7*\ch-0.5)$) -- ($(Mp)+(-0.77,-7*\ch-0.5)$);
\node[dimlabel, rotate=90] at ($(Mp)+(-1.45,-3.5*\ch)$) {$m=8$};

\draw[arrow] ($(Mp)+(4.8*\cw,-3.5*\ch)$) -- ++(0.85,0);

\coordinate (Ms) at (32.2,0);

\node[cellblue]      at ($(Ms)+(0*\cw,0*\ch)$) {1};
\node[cellfilled]    at ($(Ms)+(1*\cw,0*\ch)$) {34.95};
\node[cellblue]      at ($(Ms)+(2*\cw,0*\ch)$) {12};
\node[cellblue]      at ($(Ms)+(3*\cw,0*\ch)$) {57};
\node[cellfilled]    at ($(Ms)+(4*\cw,0*\ch)$) {46.08};

\node[cellblue]      at ($(Ms)+(0*\cw,-1*\ch)$) {32};
\node[cellblue]      at ($(Ms)+(1*\cw,-1*\ch)$) {5};
\node[cellfilled]    at ($(Ms)+(2*\cw,-1*\ch)$) {8.16};
\node[cellblue]      at ($(Ms)+(3*\cw,-1*\ch)$) {81};
\node[cellblue]      at ($(Ms)+(4*\cw,-1*\ch)$) {11};

\node[cellblue]      at ($(Ms)+(0*\cw,-2*\ch)$) {6};
\node[cellblue]      at ($(Ms)+(1*\cw,-2*\ch)$) {22};
\node[cellfilled]    at ($(Ms)+(2*\cw,-2*\ch)$) {5.48};
\node[cellblue]      at ($(Ms)+(3*\cw,-2*\ch)$) {18};
\node[cellblue]      at ($(Ms)+(4*\cw,-2*\ch)$) {23};

\node[cellblue]      at ($(Ms)+(0*\cw,-3*\ch)$) {21};
\node[cellblue]      at ($(Ms)+(1*\cw,-3*\ch)$) {17};
\node[cellfilled]    at ($(Ms)+(2*\cw,-3*\ch)$) {18.17};
\node[cellfilled]    at ($(Ms)+(3*\cw,-3*\ch)$) {213.63};
\node[cellblue]      at ($(Ms)+(4*\cw,-3*\ch)$) {29};

\node[cellfilled]    at ($(Ms)+(0*\cw,-4*\ch)$) {-27.70};
\node[cellblue]      at ($(Ms)+(1*\cw,-4*\ch)$) {31};
\node[cellblue]      at ($(Ms)+(2*\cw,-4*\ch)$) {0};
\node[cellblue]      at ($(Ms)+(3*\cw,-4*\ch)$) {19};
\node[cellblue]      at ($(Ms)+(4*\cw,-4*\ch)$) {8};

\node[cellblue]      at ($(Ms)+(0*\cw,-5*\ch)$) {31};
\node[cellfilled]    at ($(Ms)+(1*\cw,-5*\ch)$) {-18.18};
\node[cellblue]      at ($(Ms)+(2*\cw,-5*\ch)$) {18};
\node[cellblue]      at ($(Ms)+(3*\cw,-5*\ch)$) {3};
\node[cellfilled]    at ($(Ms)+(4*\cw,-5*\ch)$) {67.90};

\node[cellblue]      at ($(Ms)+(0*\cw,-6*\ch)$) {7};
\node[cellfilled]    at ($(Ms)+(1*\cw,-6*\ch)$) {20.56};
\node[cellblue]      at ($(Ms)+(2*\cw,-6*\ch)$) {13};
\node[cellfilled]    at ($(Ms)+(3*\cw,-6*\ch)$) {101.75};
\node[cellblue]      at ($(Ms)+(4*\cw,-6*\ch)$) {36};

\node[cellblue]      at ($(Ms)+(0*\cw,-7*\ch)$) {10};
\node[cellblue]      at ($(Ms)+(1*\cw,-7*\ch)$) {4};
\node[cellfilled]    at ($(Ms)+(2*\cw,-7*\ch)$) {15.61};
\node[cellblue]      at ($(Ms)+(3*\cw,-7*\ch)$) {27};
\node[cellblue]      at ($(Ms)+(4*\cw,-7*\ch)$) {61};

\node[matrixlabel] at ($(Ms)+(2*\cw,-8.65*\ch)$) {$M^{*}$};

\draw[line width=1pt] ($(Ms)+(-0.65,1.0)$) -- ($(Ms)+(4*\cw+0.65,1.0)$);
\draw[line width=1pt] ($(Ms)+(-0.65,1.0)$) -- ($(Ms)+(-0.65,0.77)$);
\draw[line width=1pt] ($(Ms)+(4*\cw+0.65,1.0)$) -- ($(Ms)+(4*\cw+0.65,0.77)$);
\node[dimlabel] at ($(Ms)+(2*\cw,1.45)$) {$n=5$};

\draw[line width=1pt] ($(Ms)+(-1,0.5)$) -- ($(Ms)+(-1,-7*\ch-0.5)$);
\draw[line width=1pt] ($(Ms)+(-1,0.5)$) -- ($(Ms)+(-0.77,0.5)$);
\draw[line width=1pt] ($(Ms)+(-1,-7*\ch-0.5)$) -- ($(Ms)+(-0.77,-7*\ch-0.5)$);
\node[dimlabel, rotate=90] at ($(Ms)+(-1.45,-3.5*\ch)$) {$m=8$};

\draw[
    -{Latex[length=4mm,width=3mm]},
    line width=1.6pt
]
($(M)+(2*\cw,1.65)$)
--
($(M)+(2*\cw,2.15)$)
--
($(Ms)+(2*\cw,2.15)$)
--
($(Ms)+(2*\cw,1.55)$);

\end{tikzpicture}

}
\caption{Illustration of matrix factorization-based missing-value reconstruction}
\label{fig:matrix_factorization}

\end{figure}

We use matrix factorization to reconstruct the missing values of \(M^{c}\) in the following way. First, we factorize the matrix \(M^{c}\) into a product of two lower-rank matrices: $M^{c} \approx F \cdot W$, where \(F\) and \(W\) are \(m \times r\) and \(r \times n^{*}\) matrices, respectively, with \(r \leq n^{*}\).  The overall factorization and reconstruction process is illustrated in Fig.~\ref{fig:matrix_factorization}. In this study, \(r=18\) was selected empirically according to preliminary reconstruction comparisons, where this setting provided the most suitable reconstruction performance for the selected variables. The implementation was based on a custom PyTorch program following the same algorithmic logic as the \texttt{MatrixFactorization} class in the \texttt{FancyImpute} Python library. After factorization, the reconstructed matrix \(M^{*}\) was obtained as follows:
\[
M^{*}[i,j] =
\begin{cases}
M^{c}[i,j], & \text{if the } ij\text{-entry } M^{c}[i,j] \text{ of } M^{c} \text{ is available,} \\
F[i,:] \times W[:,j], & \text{if } M^{c}[i,j] \text{ is missing},
\end{cases}
\]
where \(F[i,:]\) is the \(i\)-row of \(F\) and \(W[:,j]\) is the \(j\)-column of \(W\). In the Q-type models, we may calculate CNA according to the matrix \(M^{*}\) instead of \(M\) or \(M^{c}\) for each state, and set the reward function by such CNA.

\subsection{Construction of underlying MDP model: RRT-MDP}
\label{subsec3}
The Markov decision process MDP$(S,A,R,\gamma)$ framework \cite{werbos1999stable} is the underlying model of the Q-type algorithm. In this study, the MDP for RRT, denoted by RRT-MDP, is built as follows:

\begin{itemize}
\item The state space S consists of an infinite set of states, comprising the patient's indicators as depicted in Table \ref{tab:range_comparison} and the 4-dimensional RRT information. Accordingly, the state $s_t$ at each time $t$ is a 23-dimensional vector. 

\item The action space $A$ consists of three actions: $a_0$, $a_1$, and $a_2$, which denote NULL, CRRT, and IRRT actions, respectively. It should be noted that the action is defined at the level of each discrete time window rather than at the level of the entire patient trajectory. The detailed clinical interpretation of each action is provided in Table \ref{tab:action_space}.

\item The reward function R at time t is determined jointly by the CNS or SOFA score at time t and that at time t+1, as illustrated in Eq. \eqref{reward}. 

\begin{equation}
    \begin{aligned}
     &r_t^{\SOFA}= \SOFA(s_t)- \SOFA(s_{t+1}), \\
     &r_t^{\CNS}= \CNS(s_t)- \CNS(s_{t+1})
    \end{aligned}
    \label{reward}
\end{equation}

In Eq. \eqref{reward}, $\SOFA(s_t)$ denotes the SOFA score computed by the SOFA score system at state $s_t$. Similarly, CNS($s_t$) denotes the score obtained by CNA at state $s_t$, which will be defined as Eq. \eqref{CNS}. In the CNS or SOFA score, the lower the score, the better the patient's condition. Therefore, if the patients' condition improves, the score decreases, resulting in a positive reward $r_t  > 0$. Conversely, if the patient's condition worsens, a negative reward $r_t  < 0$ is obtained. However, a significant penalty is incurred if the patient dies after the treatment. The reward at the time $t$ preceding death is set to $-15$, i.e., $r_t=-15$.

\item $\gamma$ is the discount factor. We set $\gamma = 0.99$ in this study.
\end{itemize}

 \begin{table}[htbp]
     \centering
     \caption{Indicators' distribution changes with data cleaning and reconstruction}
 \label{tab:range_comparison}
 \begin{tabular}{@{}p{84pt}
 >{\centering\arraybackslash}p{52pt}
 >{\centering\arraybackslash}p{52pt}
 >{\centering\arraybackslash}p{42pt}
 >{\centering\arraybackslash}p{52pt}
 >{\centering\arraybackslash}p{42pt}
 >{\centering\arraybackslash}p{47pt}@{}}
 \toprule
    \multirow{2}{*}{\centering Indicator} & \multirow{2}{*}{\centering Ref-Vals} & \multicolumn{2}{c}{Minimum} & \multicolumn{2}{c}{Maximum} & \multirow{2}{*}{\centering $\#$Out} \\
\cmidrule(lr){3-4} \cmidrule(lr){5-6}
 & & min\textsuperscript{o} & min\textsuperscript{$\star$} & max\textsuperscript{o} & max\textsuperscript{$\star$} & \\  \hline
   \rowcolor{stdSilver!30} Age & $*$ & 18 & $*$ & 91 & $*$ & $*$ \\
   \rowcolor{stdSilver!30}   Sex & $*$ & 0 & $*$ & 1 & $*$ & $*$ \\
    \rowcolor{stdSilver!30}  Height & $*$ & 96.2 & $*$ & 203 & $*$ & $*$ \\
    \rowcolor{stdSilver!30}  Weight & $*$ & 34.5 & $*$ & 332.7 & $*$ & $*$ \\
    \rowcolor{stdSilver!30}  Vaso\_Value & $*$ & 0 & $*$ & 300 & $*$ & $*$ \\
   \rowcolor{stdSilver!30}   MV & $*$ & 0 & $*$ & 1 & $*$ & $*$ \\
   \rowcolor{stdSilver!30}   Input & $*$ & 0 & $*$ & 291938.14 & $*$ & $*$ \\
    \rowcolor{stdFuchsia!15}  Resp\_rate & $[12,20]$ & 5 & 5 & 60 & 60 & 0 \\
    \rowcolor{stdFuchsia!15} Heartrate & $[60,100]$ & 24 & 24 & 215 & 215 & 0 \\
    \rowcolor{stdFuchsia!15} Temperature & $[36,37]$ & 25.6 & 25.6 & 43.06 & 43.06 & 0 \\
    \rowcolor{stdFuchsia!15} Spo2 & $[95,100]$ & 50 & 50 & 100 & 101.11 & 51 \\
    \rowcolor{stdFuchsia!15}  Dbp & $[60,90]$ & 14 & 4.45 & 150 & 150 & 0 \\
    \rowcolor{stdFuchsia!15}  Sbp & $[90,140]$ & 40 & 40 & 215 & 215 & 0 \\
    \rowcolor{stdFuchsia!15}  Mbp & $[70,105]$ & 30 & 29.23 & 198 & 198 & 0 \\
    \rowcolor{stdFuchsia!15}  UO24 & $[1,3]$ & 0 & 0 & 100.61 & 100.61 & 0 \\
   \rowcolor{stdFuchsia!15}  Bicarbonate & $[22,27]$ & 5 & 3.15 & 40 & 40 & 0 \\
  \rowcolor{stdFuchsia!15}   Bun & $[9,20]$ & 2 & 2 & 150 & 150 & 0 \\
  \rowcolor{stdFuchsia!15}   Creatinine & $[0.5,1.5]$ & 0.1 & 0.1 & 20 & 20 & 0 \\
  \rowcolor{stdFuchsia!15}   Potassium & $[3.5,5]$ & 1.7 & 1.7 & 9 & 9 & 0 \\ 
    \toprule   
\end{tabular}
\parbox{\columnwidth}{
\vspace{4pt}
\footnotesize \justifying \noindent
The min\textsuperscript{o} and max\textsuperscript{o} denote the minimum and maximum in the original dataset $M$, respectively. The min\textsuperscript{$\star$} and max\textsuperscript{$\star$} denote the minimum and maximum in the reconstruction dataset $M^{\star}$, respectively. The $\#$Out denotes the number of a reconstructed indicator out of the range $[\text{min}^{\text{o}},\text{max}^{\text{o}}]$. The symbol $*$ denotes no data reconstruction for the corresponding indicator. There are only 51 reconstructed indicators (with the ratio $2\times10^{-5}$ to the number of reconstructed values 152,753) out of the range $[\text{min}^{\text{o}},\text{max}^{\text{o}}]$. After imputation, only the reconstructed values of Spo2 exceeded the original value range, containing just 51 outliers, accounting for 0.033\% of the total dataset.
}
 \end{table}

\begin{table}[htbp]
\centering
\caption{Clinical interpretation of the action space in the RRT-MDP}
\label{tab:action_space}
\resizebox{\textwidth}{!}{
\begin{tabular}{lll}
\hline
Action & Treatment status & Clinical interpretation \\
\hline
$a_0$ & NULL & No RRT intervention was recorded within the current time window. \\
$a_1$ & CRRT & Continuous renal replacement therapy was delivered or ongoing within the current time window. \\
$a_2$ & IRRT & Intermittent renal replacement therapy was delivered within the current time window. \\
\hline
\end{tabular}
}
\end{table}

\subsection{CNA and its derived reward function in Q-type algorithms}
\label{subsec4}
Given a vital sign with a continuous value, its value has two types of distribution. One is its normal distribution for healthy people, called the referent interval $[l, u]$, where $l$ and $u$ are the lower and upper bounds, respectively, fixed by medical investigations. Another distribution is for live people, called the limit interval $[min, max]$, where $min$ and $max$ are the lower and upper bounds, respectively, counted from the available data for that vital sign. The values of a patients’ vital signs may deviate from the referent interval $[l, u]$ from two directions. When $max-u \neq l-min$, although $d_2 - max = min - d_1$, the deviations $d_1$ and $d_2$ ought to indicate the different health status of a patient. For instance, the referent interval of SBP is $[90, 140]$, and the SBP $70$ is usually thought to be worse than SBP $160$.

If any endpoint value of $l$ and $u$ is negative, we need to take a shift to make the values $l, u, min$, and $max$ non-negative, as follows:
\begin{equation}
l'= l + min, u' = u + min, {min}' = 0, {max}' = max + min   
\end{equation}
After this, or the limit values are inherently non-negative, we utilize power function $x^p$ to transform the values
$l,u,min, \text{and}\ max$ into
\begin{equation}
    l' = l^p, \quad u' = u^p, \quad {min}' = {min}^p, \quad {max}' = {max}^p
\end{equation}
such that
\begin{equation}
    u' - {max}' = {min}' - l'
\end{equation}
By mathematical theory, such $p$ always exists for arbitrary nonnegative $l,u,min$, and $max$.

\begin{sloppypar}
Suppose two vital signs $s_1$ and $s_2$ have the intervals $[l_1, u_1]$ and $[l_2, u_2]$, and limit intervals $[min_1, max_1]$ and $[min_2, max_2]$ such that $u_1 - max_1 = min_1 - l_1$ and $u_2 - max_2 = min_2 - l_2$ but $u_1 - max_1 \neq u_2 - max_2$. In such cases, the values $v_{s_1}$ and $v_{s_2}$ of $s_1$ and $s_2$ with the same deviation $d = u_1 - v_{s_1} = u_2 - v_{s_2}$ indicate different degrees of deviation. To address this issue, we utilize the three-sigma rule of the normal distribution to standardize the relationship between the reference interval $[l', u']$ and the limit interval $[min', max']$ as follows.
\end{sloppypar}

First, we take a shift
\begin{equation}
\begin{aligned}
l^s &= l' - \frac{l' + u'}{2}, &\quad u^s &= u' - \frac{l' + u'}{2}, \\
{min}^s &= {min}' - \frac{l' + u'}{2}, &\quad {max}^s &= {max}' - \frac{l' + u'}{2}
\end{aligned}
\end{equation}
to make the center of these intervals at the original point.
Then, we utilize the hyperbolic sine function \(\sinh_\alpha(x) = \frac{e^{\alpha x} - e^{-\alpha x}}{2}\) and \(\tanh_\alpha(x) = \frac{e^{\alpha x} - e^{-\alpha x}}{e^{\alpha x} + e^{-\alpha x}}\), with parameter $\alpha$. If \({max}^s > 3u^s\),  we adjust \(l^s, u^s, {min}^s, {max}^s\) to
\begin{equation}
\begin{aligned}
l^\star &= \tanh_\alpha(l^s), \quad & u^\star &= \tanh_\alpha(u^s), \\
{min}^\star &= \tanh_\alpha({min}^s), \quad & {max}^\star &= \tanh_\alpha({max}^s),
\end{aligned}
\end{equation}
else, we adjust \(l^s, u^s, {min}^s, {max}^s\) to
\begin{equation}
\begin{aligned}
l^\star &= \sinh_\alpha(l^s), \quad & u^\star &= \sinh_\alpha(u^s), \\
{min}^\star &= \sinh_\alpha({min}^s), \quad & {max}^\star &= \sinh_\alpha({max}^s),
\end{aligned}
\end{equation}
such that \(l^\star, u^\star, {min}^\star, {max}^\star\) are samples of a normal distribution \(N(0, \sigma)\) with \(\sigma = -l^\star = u^\star\), and \({max}^\star = -{min}^\star = 3u^\star\).

At last, we standardize the normal distribution \(N(0, \sigma)\) to \(N(0, 1)\) by setting
\begin{equation}
l^* = \frac{l^\star}{u^\star}, \quad u^* = 1, \quad min^* = \frac{min^\star}{u^\star}, \quad max^* = \frac{max^\star}{u^\star}
\end{equation}

In summary, given a value $v$ of a vital sign with the referent interval $[l,u]$ and the limit interval $[min,max]$, $v$ is transformed into a value $v^*$ by the following equation
\begin{equation}
   v^* = \frac{1}{\sinh_\alpha\!\left(\frac{u^p - l^p}{2}\right)} \sinh_\alpha\!\left(v^p - \frac{u^p + l^p}{2}\right)
\end{equation}
or
\begin{equation}
    v^* = \frac{1}{\tanh_\alpha\!\left(\frac{u^p - l^p}{2}\right)} \tanh_\alpha\!\left(v^p - \frac{u^p + l^p}{2}\right)
\end{equation}
Using such $v^*$, we establish a uniform metric on all continuous vital signs in Table \ref{tab:range_comparison}. Given dataset {$v$}, the corresponding set {$v^*$} is called the normalized dataset of {$v$}. The operational workflow of CNA/CNS is summarized in Algorithm~\ref{alg:cna_cns}.

\begin{algorithm}[!htbp]
\caption{Computation process of CNA and CNS.}
\label{alg:cna_cns}
\scriptsize

\begin{algorithmic}[1]

\State $\{v_i\}_{i=1}^{K},\{[l_i,u_i]\}_{i=1}^{K},\{[min_i,max_i]\}_{i=1}^{K}\leftarrow InputValues$, $V^*\leftarrow []$

\Statex \hspace{\algorithmicindent}\textbf{Step 1: Normalize each vital sign interval}

\For{$i\leftarrow 1$ to $K$}

    \If{$\min(l_i,u_i,min_i,max_i)<0$}
        \State $l_i'\leftarrow l_i+min_i$, $u_i'\leftarrow u_i+min_i$, $min_i'\leftarrow0$, $max_i'\leftarrow max_i+min_i$, $v_i'\leftarrow v_i+min_i$
    \Else
        \State $l_i'\leftarrow l_i$, $u_i'\leftarrow u_i$, $min_i'\leftarrow min_i$, $max_i'\leftarrow max_i$, $v_i'\leftarrow v_i$
    \EndIf

    \State Find $p$ satisfying the required symmetry condition
    \State $l_i^p\leftarrow(l_i')^p$, $u_i^p\leftarrow(u_i')^p$, $min_i^p\leftarrow(min_i')^p$, $max_i^p\leftarrow(max_i')^p$, $v_i^p\leftarrow(v_i')^p$

    \State $c_i\leftarrow(l_i^p+u_i^p)/2$
    \State $l_i^s\leftarrow l_i^p-c_i$, $u_i^s\leftarrow u_i^p-c_i$, $min_i^s\leftarrow min_i^p-c_i$, $max_i^s\leftarrow max_i^p-c_i$, $v_i^s\leftarrow v_i^p-c_i$

    \Statex \hspace{\algorithmicindent}\textbf{Step 2: Apply nonlinear mapping}

    \If{$max_i^s>3u_i^s$}
        \State Choose $\alpha$ satisfying the three-sigma rule
        \State $l_i^\star\leftarrow\tanh_\alpha(l_i^s)$, $u_i^\star\leftarrow\tanh_\alpha(u_i^s)$, $min_i^\star\leftarrow\tanh_\alpha(min_i^s)$, $max_i^\star\leftarrow\tanh_\alpha(max_i^s)$, $v_i^\star\leftarrow\tanh_\alpha(v_i^s)$
    \Else
        \State Choose $\alpha$ satisfying the three-sigma rule
        \State $l_i^\star\leftarrow\sinh_\alpha(l_i^s)$, $u_i^\star\leftarrow\sinh_\alpha(u_i^s)$, $min_i^\star\leftarrow\sinh_\alpha(min_i^s)$, $max_i^\star\leftarrow\sinh_\alpha(max_i^s)$, $v_i^\star\leftarrow\sinh_\alpha(v_i^s)$
    \EndIf

    \Statex \hspace{\algorithmicindent}\textbf{Step 3: Standardize the transformed values}

    \State $l_i^*\leftarrow l_i^\star/u_i^\star$, $u_i^*\leftarrow1$, $min_i^*\leftarrow min_i^\star/u_i^\star$, $max_i^*\leftarrow max_i^\star/u_i^\star$, $v_i^*\leftarrow v_i^\star/u_i^\star$
    \State $V^*.append(v_i^*)$

\EndFor

\Statex \hspace{\algorithmicindent}\textbf{Step 4: Identify abnormal normalized values}

\State $S\leftarrow []$

\For{$i\leftarrow1$ to $K$}
    \If{$v_i^*<-1$ or $v_i^*>1$}
        \State $S.append(i)$
    \EndIf
\EndFor

\Statex \hspace{\algorithmicindent}\textbf{Step 5: Compute CNA components}

\State $d_r\leftarrow |S|/K$
\State $d_m\leftarrow Mean(\{|v_i^*|:i\in S\})$ if $|S|>0$, else $0$
\State $d_e\leftarrow Variance(\{v_i^*:i\in S\})$ if $|S|>1$, else $0$

\Statex \hspace{\algorithmicindent}\textbf{Step 6: Compute CNS score}

\State $CNS\leftarrow0.6d_r+0.3d_m+0.1d_e$
\State \Return $(d_r,d_m,d_e)$ and $CNS$

\end{algorithmic}
\end{algorithm}

Now, we face the problem of summarizing the ensemble of all deviations of vital signs’ values. A naive way is to add all deviations. However, such a simple summation can’t distinguish between healthy and ill conditions. For instance, a person with $v^*$ being $-2,-2,2,2$ on some vital signs is sick, but a person with $v^*$ being 0,0,0,0 on the same vital signs is in healthy status. Notice that the means of those two cases are the same. Thus, the mean value of $v^*$ also cannot discriminate the situations. To address this issue,  we use the distance between the normalized value $v^*$ and the reference interval $[-1,1]$ to quantify the deviation degree of each vital sign.  However, this is still insufficient to distinguish the situations such as that between the deviations $0.1,0.1,0.1,0.1,0.1,0.1,0.1,0.1$ of $-1.1,-1.1,-1.1,-1.1,1.1,1.1,1.1,1.1$ and $1,0,0,0,0,0,0,0,0,0$ of $-2,-1,-1,-1,1,1,1,1$. Against such cases, we introduce a new metric vector ($d_r,d_m,d_e$), called the comprehensive normalized assessment (CNA).  Specifically, $d_r$ represents the ratio of abnormal indicators, namely the proportion of vital signs outside their reference intervals; $d_m$ represents the average degree of abnormality, measured by the mean absolute deviation of the abnormal normalized values; and $d_e$ represents the dispersion of abnormality, measured by the variance of the abnormal normalized values. Each component is nonnegative and directly derived from deviations from clinically defined reference intervals. In this sense, CNA provides an interpretable assessment at the health-status metric level by decomposing a complex patient state into three clinically readable components: how many indicators are abnormal, how abnormal they are on average, and how dispersed the abnormalities are across indicators. It is noteworthy that each of $d_r$,$d_m$, and $d_e$ is nonnegative.

Although the CNA vector \((d_r,d_m,d_e)\) provides an interpretable multidimensional assessment of health status, it cannot be directly used as a scalar reward in Q-learning-based algorithms. Therefore, we further summarize these components into the Comprehensive Normalized Score (CNS) using a weighted summation:
\begin{equation}
    \CNS(v_s)=0.6d_r+0.3d_m+0.1d_e,
    \label{CNS}
\end{equation}
 where \(v_s\) denotes the set of vital signs used to calculate \(d_r\), \(d_m\), and \(d_e\). The weights were determined according to the relative associations between the three CNA components and mortality, estimated using statistical testing. The resulting CNS is a scalar health-status score, where a lower value indicates a better physiological condition. In the reinforcement learning framework, the temporal change in CNS between two adjacent states is used to construct the reward function, thereby providing a continuous and fine-grained learning signal for strategy optimization.

\begin{table}[htbp]
\caption{Treatment Effect Comparison Matrix (TECM)}
\centering
\label{tab:TECM}
\begin{tabular}{@{}lcc@{}}
\toprule
 & \textbf{Optimal strategy: $\pi^*$} & \textbf{Worst strategy: $\pi'$} \\
\midrule
\textbf{``Good'' actual treatments: $P_G$} & OG & WG \\
\textbf{``Bad'' actual treatments: $P_B$} & OB & WB \\
\bottomrule
\end{tabular}
\parbox{0.85\columnwidth}{
\vspace{4pt}
\footnotesize \justifying \noindent
$P_G$, ``good'' physicians' actual treatments; $P_B$, ``bad'' physicians' actual treatments; OG, a consistency measure between the optimal strategy and the ``good'' physicians' actual treatments; OB, a disagreement measure between the optimal strategy and the ``bad'' physicians' actual treatments; WG, a disagreement measure between the worst strategy and the ``good'' physicians' actual treatments; WB, a consistency measure between the worst strategy and the ``bad'' physicians' actual treatments.
}
\end{table}
\subsection{Data-driven concept of good/bad treatment}
\label{subsec5}
In the analyses of the performances of AI strategies, we adopted a recent data-driven assessment method TECM* in \cite{liu2025tecm}, which consists of a Treatment Effect Comparison Matrix (TECM) defined as Table \ref{tab:TECM}, O-gap $\Xi^o := \mathrm{OG} - \mathrm{OB}$, W-gap $\Xi^w := \mathrm{WB} - \mathrm{WG}$, comprehensive confidence $\sigma$ and comprehensive bias $\mu$ defined as Eq. \eqref{sigma_mu}.
\begin{equation}
\sigma = \frac{2 \cdot \mathrm{OG} \cdot \mathrm{WB}}{(\mathrm{OG} + \mathrm{WB})} \cdot \frac{(\mathrm{OG} + \mathrm{WG})}{(2 \cdot \mathrm{OG} \cdot \mathrm{WG})}, \quad \mu = (\mathrm{OG} - \mathrm{WG}) - (\mathrm{WB} - \mathrm{OB})
\label{sigma_mu}
\end{equation}

The ``good'' and ``bad'' physicians' actual treatments in Table \ref{tab:TECM} are defined as follows. Given a physicians' actual treatment $ep = (s_0,a_0; ...; s_{n-1}  = s_{ter},\phi)$. Let $N_g=|{t:r_t\ge0}|$, where $r_t$ is a reward function like $r_t^{SOFA}$ or $r_t^{CNS}$, and the subscript $g$ stands for ``good''. The effective rate $\rho_{ep}$ of this episode ep is defined as $\rho_{ep}=N_g/n$. Given a set P of episodes and a threshold $\tau$, if $\rho_{ep}\ge\tau$, we classify this ep as a ``good'' physicians' actual treatment and add it to the set $P_G$. Otherwise, it is classified as a ``bad'' physicians' actual treatment and added to the set $P_B$.

\subsection{Novel assessment metrics: TECM*}
\label{subsec6}
Assume \( Q : (S, A) \mapsto \mathbb{R} \) is the value function learned by an AI method. Let \(\pi^*\) and \(\pi'\) denote the best and worst AI strategy, respectively, which are defined as  
\begin{equation}
   \pi^*(s) = \arg\max_{a \in A} Q(s, a), \quad \pi'(s) = \arg\min_{a \in A} Q(s, a). 
\end{equation}
for each state $s \in S$.

It is worth noting that the evaluation of \(Q(s,a)\) based on merely the optimal strategy \(\pi^*\) derived by \(Q(s,a)\) may lose some information inside \(Q(s,a)\). For a comprehensive evaluation of \(Q(s,a)\), we consider the worst AI strategy \(\pi'\) also derived by \(Q(s,a)\). Within the set \(P_G\) and the set \(P_B\), we compute the optimal strategy \(\pi^*(s_t)\) and the worst strategy \(\pi'(s_t)\) for each state \(s_t\) in each episode$\mathrm{ep} = (s_0, a_0; s_1, a_1; \dots; s_{n-1}=s_{\text{ter}}, a_{n-1}=\varnothing)$. The similarity between the actual treatment action \(a_t\) and the optimal strategy \(\pi^*(s_t)\), denoted as $\mathrm{sim}_o(\mathrm{ep},t) = \frac{1}{1 + 0.25 \cdot |a_t - \pi^*(s_t)|}$, was calculated for every time step, as was the similarity to the action recommended by the worst strategy \(\pi'\), denoted as $\mathrm{sim}_w(\mathrm{ep},t) = \frac{1}{1 + 0.25 \cdot |a_t - \pi'(s_t)|}$.

For each $\mathrm{ep} \in P_G$, we calculated:
\begin{equation}
    N_{og}(\mathrm{ep}) = |\{t : \mathrm{sim}_o(\mathrm{ep},t) \geq \mathrm{sim}_w(\mathrm{ep},t)\}| 
\end{equation}
\begin{equation}
    \rho_{og} = \frac{N_{og}(\mathrm{ep})}{n}
\end{equation}
\begin{equation}
    \rho_{wg} = 1 - \rho_{og}
\end{equation}
If $\rho_{og} > \tau$, then $\mathrm{ep} \in \text{opt-pol}(P_G)$; otherwise, $\mathrm{ep} \in \text{wrt-pol}(P_G)$. The average similarity
\[
\begin{aligned}
\mathrm{OG} &= \frac{\sum{\mathrm{ep}\in \mathrm{opt}\text{-}\mathrm{pol}(P_G)} \frac{\sum_t \mathrm{sim}_o(\mathrm{ep},t)}{|\mathrm{ep}|}}{|\mathrm{opt}\text{-}\mathrm{pol}(P_G)|} \cdot \rho_{og}, \\
\text{and } \mathrm{WG} &= \frac{\sum_{\mathrm{ep} \in \text{wrt-pol}(P_G)} \frac{\sum_t \mathrm{sim}_w(\mathrm{ep},t)}{|\mathrm{ep}|}}{|\text{wrt-pol}(P_G)|} \cdot \rho_{wg}
\end{aligned}
\]
were computed for $\text{opt-pol}(P_G)$ and $\text{wrt-pol}(P_G)$, respectively.

For each $\mathrm{ep} \in P_B$, we calculated:
\begin{equation}
    N_{wb}(\mathrm{ep}) = |\{t : \mathrm{sim}_w(\mathrm{ep},t) \geq \mathrm{sim}_o(\mathrm{ep},t)\}|
\end{equation}
\begin{equation}
    \rho_{wb} = \frac{N_{wb}}{n}   
\end{equation}
\begin{equation}
    \rho_{ob} = 1 - \rho_{wb}
\end{equation}
If $\rho_{ob} > \tau$, $\mathrm{ep} \in \text{opt-pol}(P_B)$; otherwise, $\mathrm{ep} \in \text{wrt-pol}(P_B)$. The average similarities
\[
\begin{aligned}
\mathrm{OB} &= \frac{\sum_{\mathrm{ep} \in \text{opt-pol}(P_B)} \frac{\sum_t \mathrm{sim}_o(\mathrm{ep},t)}{|\mathrm{ep}|}}{|\text{opt-pol}(P_B)|} \cdot \rho_{ob}, \\
\text{and } \mathrm{WB} &= \frac{\sum_{\mathrm{ep} \in \text{wrt-pol}(P_B)} \frac{\sum_t \mathrm{sim}_w(\mathrm{ep},t)}{|\mathrm{ep}|}}{|\text{wrt-pol}(P_B)|} \cdot \rho_{wb}
\end{aligned}
\]
were computed for $\text{opt-pol}(P_B)$ and $\text{wrt-pol}(P_B)$, respectively.

\subsection{Strategy selection via TECM*-principle}
\label{subsec7}
Determining the optimal point to terminate the training process or establish convergence of Q-type algorithms remains a significant challenge, especially for models operating in continuous state spaces. To address this issue, we leverage the data-driven \(\eta\)-TECM* termination criterion established in \cite{werbos1999stable}. This termination criterion is built on the following strategy assessment method. Comparing the O-gaps \(\Xi^o(\pi_i)\) and W-gaps \(\Xi^w(\pi_i)\) of two strategies \(\pi_i\) for \(i = 1,2\), the optimal strategy is determined by the following three rules:
\begin{itemize}
    \item If \(\Xi^o(\pi_i) \geq \Xi^o(\pi_j)\), and \(\Xi^w(\pi_i) \geq \Xi^w(\pi_j)\), for \(i \neq j \in [1,2]\), then \(\pi_i\) is said to be better than \(\pi_j\).
    \item Otherwise, if \(\Xi^o(\pi_i) \geq \Xi^o(\pi_j)\) and \(\Xi^w(\pi_i) < \Xi^w(\pi_j)\) for \(i \neq j \in [1,2]\), the optimal strategy is determined by comprehensive confidence \(\sigma\) or comprehensive bias \(\mu\). As \(\sigma\) is used, \(\pi_i\) is said to be better than \(\pi_j\) if \(\sigma(\pi_i) \geq \sigma(\pi_j)\).
    \item When \(\mu\) is used in the second case, assume \(\sigma(\pi_i) \geq \sigma(\pi_j)\), if a conservative strategy is chosen then \(\pi_j\) is said to be better than \(\pi_i\), otherwise \(\pi_i\) is the better one. 
\end{itemize}

The \(\eta\)-TECM* termination criterion works as follows. Once the best strategy is not chosen by the above three-rule principle on the \(\eta\) highest epochs, the algorithm should terminate, where \(\eta \geq 1\) is a hyperparameter. In this study, we picked \(\eta = 50\).

\section{Results}
\subsection{Data preprocessing result}
The vital signs in Table \ref{tab:missing_rates} span multiple medical scoring systems. Using any single score for reward function design may inadequately represent patient health status. Furthermore, these signs exhibit substantial missing values in both MIMIC-IV and eICU databases (shown in Table \ref{tab:missing_rates}). Incomplete data is common in multi-source datasets. For instance, Glucose values are missing in 105,308 MIMIC-IV records versus 43,632 in eICU. Since discarding incomplete records would drastically reduce sample size, we utilize records with limited missingness provided key variables are available or overall missingness remains acceptable.

As the vital signs are indicators of a life system, they should be related. When the missing values of a sign are limited, they may be reconstructed by other existing data. In this study, postulating such reciprocity among vital signs, we explored matrix factorization for data reconstruction \cite{nguyen2019low}. To capture the interactions among vital signs, we hold the following \(\theta\)-majority principle for cleaning data with missing values: Any vital sign with a missing rate bigger than \(\theta\) will be cleaned. In other words, we only reconstruct the values for the vital signs that their missing rates are less than \(\theta\). Moreover, we will calculate the CNS based on such vital signs to set reward functions.

This study employed matrix factorization with \(\theta\)-majority principle (\(\theta = 40\%\)) for vital signs reconstruction. Initial feature selection yielded only 7 indicators, insufficient to cover clinically significant features for RRT treatment per Standard Operating Procedures. Consequently, \(\theta\) was relaxed to \(70\%\) to expand the feature set. Although matrix factorization is methodologically unsuitable for reconstructing static attributes and therapeutic interventions, these metrics were included to enhance overall reconstruction accuracy while retaining original values in the final output. The final set of 23 indicators formed an \(n \times 23\) matrix \(M^c\) containing 152,753 missing values (\(24.71\%\) missing rate). Factorization \(M^c \approx F \cdot W\) achieved a loss of 0.53, with reconstruction effects on data distribution shown in Table \ref{tab:range_comparison}.

\subsection{Performances of models}
To assess the performance of those algorithms, we adopt the following two assessment methods. One is the assessment method in \cite{werbos1999stable} based on the statistical comparison with historical treatments (SCwHT for short). Accordingly, SCwHT checks whether AI strategy could reduce the mortality rate or length of ICU stay by comparing episodes in the test dataset that followed AI strategy with those that did not. For research justification, it compares the group that followed AI strategy by matching the action in the episode. If the action of the physicians’ actual treatment matched the AI strategy recommendation and its period comprised more than 65\% of the whole data period of the episode, then it will be considered as an AI strategy follower.

\subsection{Performances of models with MF}
\begin{figure}[!htbp]
    \centering
{ \footnotesize
\begin{tabular}{cl cl cl cl}
{ \rule[0.5ex]{0.5cm}{0.8pt}} & DQN-MF-CNS & { \rule[0.5ex]{0.5cm}{0.8pt}} & DDQN-MF-CNS & { \rule[0.5ex]{0.5cm}{0.8pt}} & BCQ-MF-CNS & { \rule[0.5ex]{0.5cm}{0.8pt}} & CQL-MF-CNS \\
{ \rule[0.5ex]{0.2cm}{0.8pt} \rule[0.5ex]{0.2cm}{0.8pt}} & DQN-MF-SOFA & { \rule[0.5ex]{0.2cm}{0.8pt} \rule[0.5ex]{0.2cm}{0.8pt}} & DDQN-MF-SOFA & { \rule[0.5ex]{0.2cm}{0.8pt} \rule[0.5ex]{0.2cm}{0.8pt}} & BCQ-MF-SOFA & { \rule[0.5ex]{0.2cm}{0.8pt} \rule[0.5ex]{0.2cm}{0.8pt}} & CQL-MF-SOFA  
\end{tabular}
}
    \SubFigTL{0.48\textwidth}{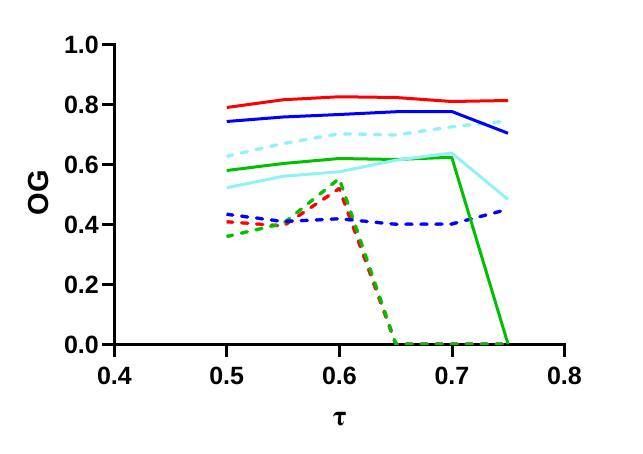}{a}{fig:tecm_a}
    \hfill
    \SubFigTL{0.48\textwidth}{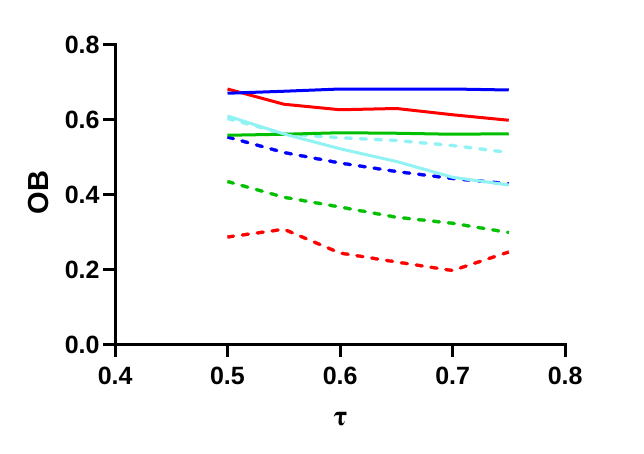}{b}{fig:tecm_b}
    \SubFigTL{0.48\textwidth}{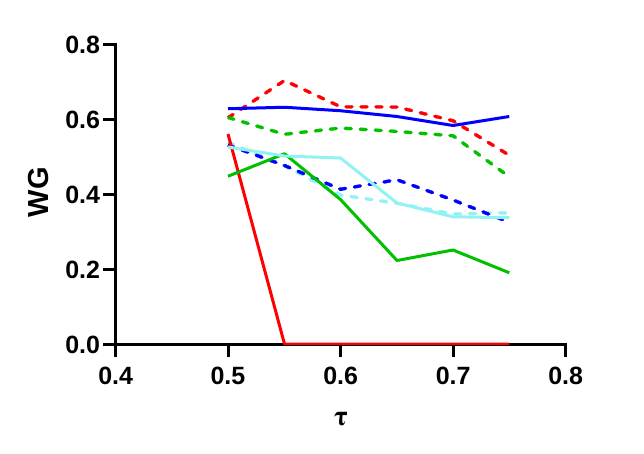}{c}{fig:tecm_c}
    \hfill
    \SubFigTL{0.48\textwidth}{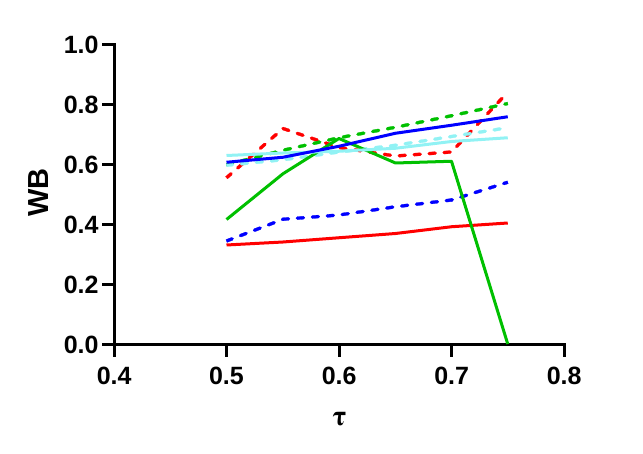}{d}{fig:tecm_d}
    \vspace{4mm}
    \SubFigTL{0.48\textwidth}{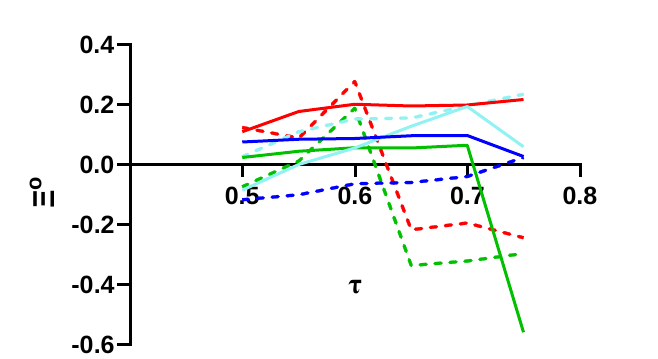}{e}{fig:tecm_e}
    \hfill
    \SubFigTL{0.48\textwidth}{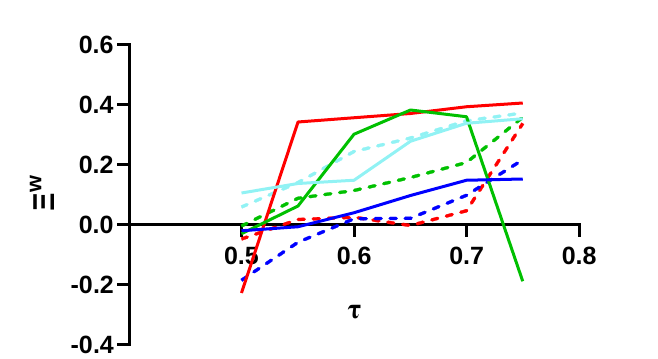}{f}{fig:tecm_f}
    \SubFigTL{0.48\textwidth}{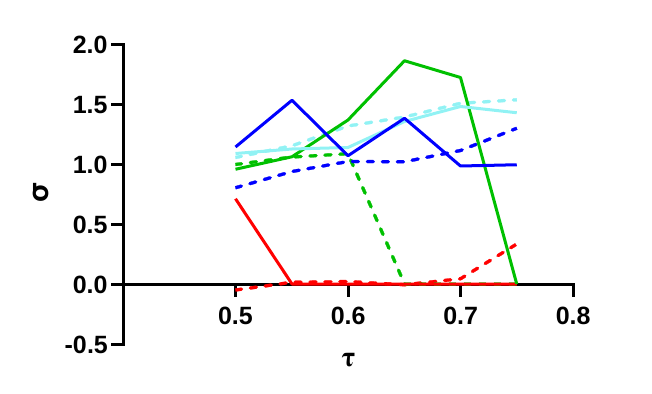}{g}{fig:tecm_g}
    \hfill
    \SubFigTL{0.48\textwidth}{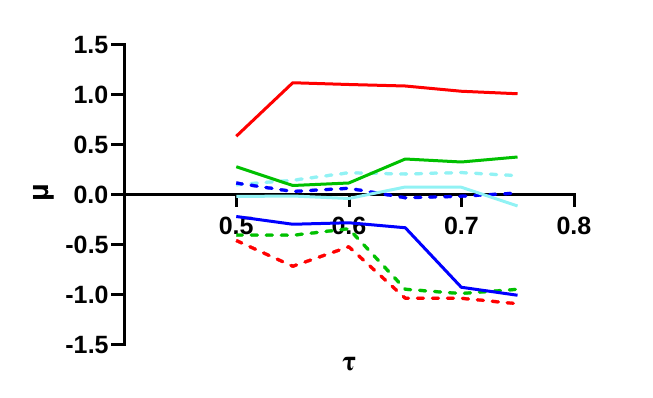}{h}{fig:tecm_h}
    \caption{TECM* of models via action similarity rate. Subfigure a-h represent the results for OG, OB, WG, WB, O-Gap $\Xi^o$, W-Gap $\Xi^w$, comprehensive confidence $\sigma$, and comprehensive bias $\mu$ of eight models}
    \label{fig:TECM_diffmodel}
\end{figure}

Using Eq. \eqref{reward} to quantify rewards following SOFA computation, we implemented four DRL models (DQN, DDQN, BCQ, CQL) designated as SOFA-DQN series. Corresponding MF-SOFA series models were developed using matrix factorization-reconstructed data. Training employed \(\eta\)-TECM* termination criterion (\(\eta = 50\)) with action similarity rate \cite{liu2024value} as the basic metric. Similarly, we evaluate performances of those eight models for the hyperparameter \(\tau \in [0.5, 0.75]\). The results presented in Fig. \ref{fig:TECM_diffmodel} demonstrate that the optimal AI strategies improve the physicians' actual treatment, and each MF-SOFA version model achieves substantially superior performance compared to models that directly used the SOFA. Moreover, among those models, the optimal AI strategy of CQL-MF-SOFA offers the best fitting around \(\tau = 0.65\).

Similarly, we compute those strategies’ mortality rates and length of ICU stays on the external testing dataset, by the comparison method in \cite{werbos1999stable}. All results in this table were also subjected to the \(p\)-value \(< 0.05\), indicating them statistically significant. The results are reported in the last two rows of Table \ref{tab:performances_compact}. Among these AI strategies, CQL-MF-SOFA has the best performance, which improves the mortality rate from 13.24\% to 7.12\% (reducing 46.22\%) and decreases the average length of ICU stay from 308.48 to 266.35 hours (reducing \(13.66\%\)).
\begin{table}[htbp]
\centering
\caption{Performances of MF-SOFA \& SOFA models with $\tau = 0.65$}
\label{tab:performances_compact}
\small
\resizebox{\textwidth}{!}{
\begin{tabular}{@{}lccccccccc@{}}
\toprule
\multirow{2}{*}{\centering Metric} & 
\multirow{2}{*}{\centering PAT} & 
\multicolumn{2}{c}{DQN-} & 
\multicolumn{2}{c}{DDQN-} & 
\multicolumn{2}{c}{BCQ-} & 
\multicolumn{2}{c}{CQL-} \\
\cmidrule(lr){3-4} \cmidrule(lr){5-6} \cmidrule(lr){7-8} \cmidrule(lr){9-10}
 & & MF & SOFA & MF & SOFA & MF & SOFA & MF & SOFA \\
\midrule
OG & $\times$ & 0.636 & 0.540 & 0.583 & 0.570 & 0.482 & 0.463 & 0.635 & 0.635 \\
OB & $\times$ & 0.471 & 0.511 & 0.450 & 0.460 & 0.370 & 0.361 & 0.477 & 0.486 \\
WG & $\times$ & 0.562 & 0.688 & 0.286 & 0.670 & 0.626 & 0.577 & 0.381 & 0.434 \\
WB & $\times$ & 0.637 & 0.761 & 0.736 & 0.754 & 0.716 & 0.685 & 0.599 & 0.617 \\
$\sigma$ & $\times$ & 1.06 & 1.043 & 1.694 & 1.047 & 1.057 & 1.075 & 1.293 & 1.213 \\
$\mu$ & $\times$ & $-0.091$ & $-0.397$ & 0.011 & $-0.403$ & $-0.489$ & $-0.438$ & 0.131 & 0.070 \\
\rowcolor{yellow!35}
BE & $\times$ & 500 & 300 & 400 & 200 & 400 & 400 & 400 & 200 \\
\rowcolor{green!35}
MoR & 13.24 & 9.76 & 10.00 & 9.16 & 9.26 & 7.52 & 8.86 & \textbf{7.12} & 8.11 \\
\rowcolor{green!35}
AiHS & 308.48 & 279.09 & 294.65 & 275.83 & 284.43 & 274.60 & 270.92 & \textbf{266.35} & 269.23 \\
\bottomrule
\end{tabular}
}
\parbox{0.98\columnwidth}{
\vspace{4pt}
\footnotesize \raggedright
PAT: Physicians' actual treatment; BE: Best epoch; MoR: Mortality rate; AiHS: Average ICU stay. 
}
\end{table}

\subsection{Comparison of SOFA with CNS models}
Under the framework \ref{fig:state_vital}, with data reconstruction, we developed four MF-CNS models, called the associated models accordingly the DQN-MF-CNS, DDQN-MF-CNS, BCQ-MF-CNS, and CQL-MF-CNS. To evaluate the optimization effect of CNS, we established corresponding MF-SOFA models using SOFA score as baseline, selected for its superior prognostic prediction per reference \cite{oh2021optimal}.

\begin{figure}[!htbp]
    \centering
    \SubFigTL{0.48\textwidth}{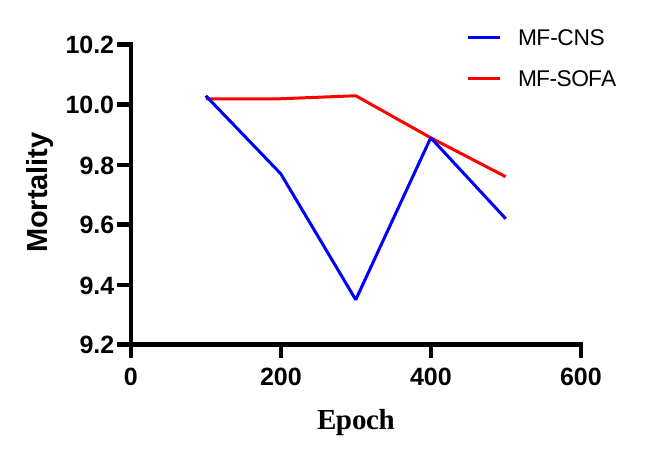}{a}{fig:Mortality_a}
    \hfill
    \SubFigTL{0.48\textwidth}{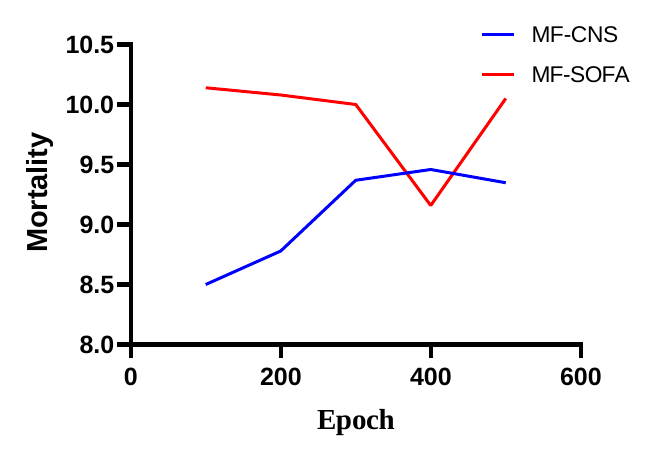}{b}{fig:Mortality_b}
    \SubFigTL{0.48\textwidth}{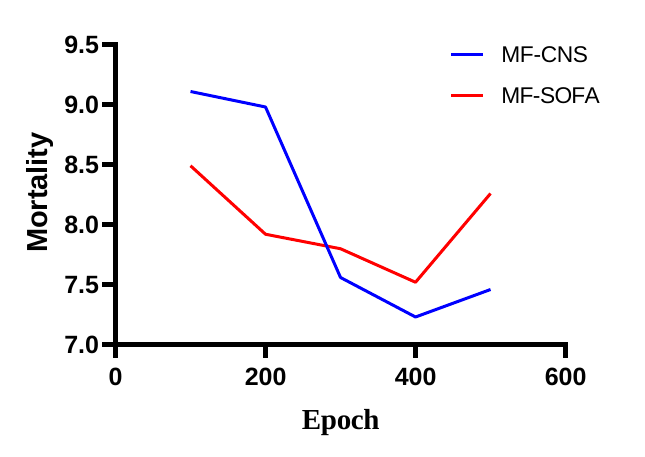}{c}{fig:Mortality_c}
    \hfill
    \SubFigTL{0.48\textwidth}{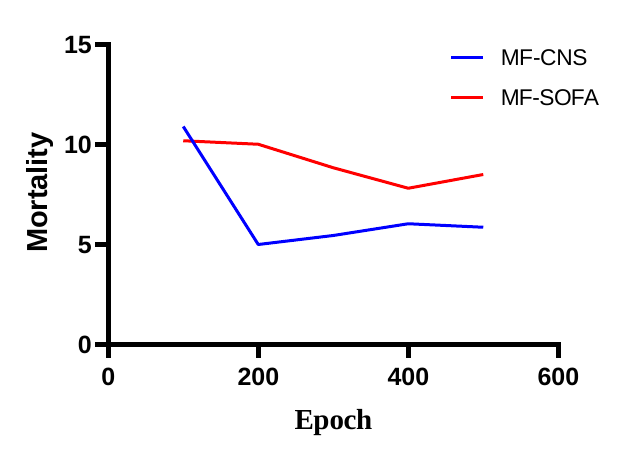}{d}{fig:Mortality_d}
    \caption{Mortality rates with MF-SOFA \& MF-CNS. Subfigure a represents the mortality rate corresponding to DQN; Subfigure b represents the mortality rate corresponding with DDQN; Subfigure c represents the mortality rate corresponding with BCQ; Subfigure d represents the mortality rate corresponding with CQL}
    \label{fig:Mortality}
\end{figure}

\begin{figure}[!htbp]
	\centering
	\SubFigTL{0.48\textwidth}{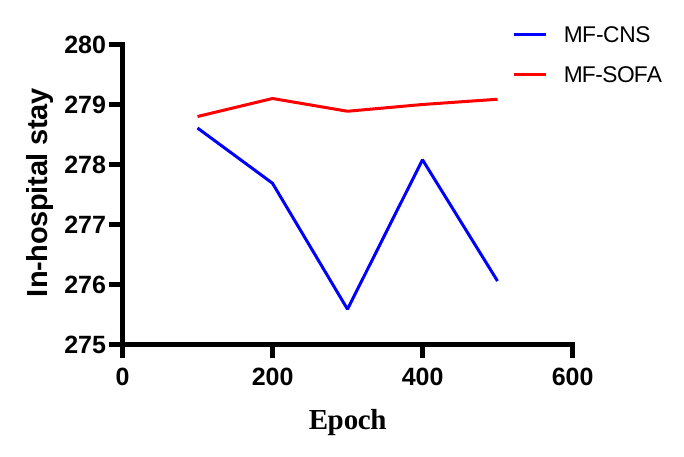}{a}{fig:hospital_a}
	\hfill
	\SubFigTL{0.48\textwidth}{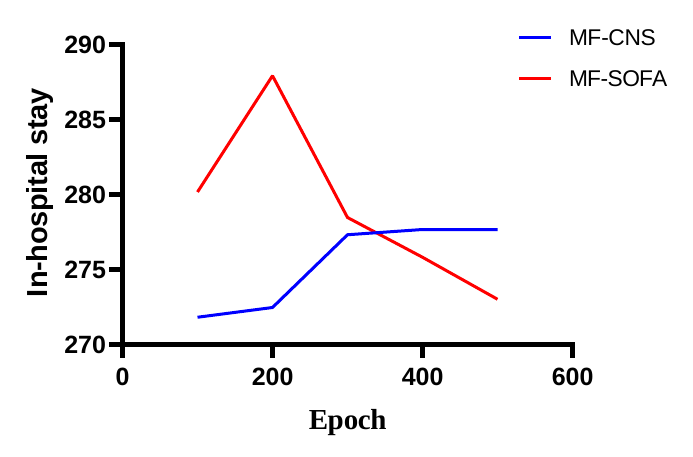}{b}{fig:hospital_b}
	\SubFigTL{0.48\textwidth}{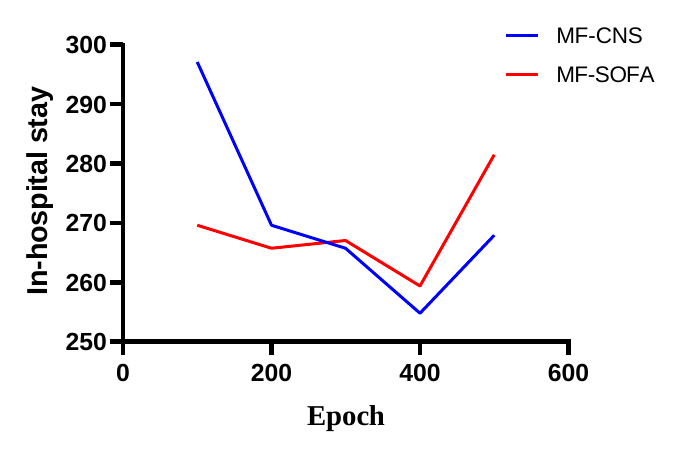}{c}{fig:hospital_c}
	\hfill
	\SubFigTL{0.48\textwidth}{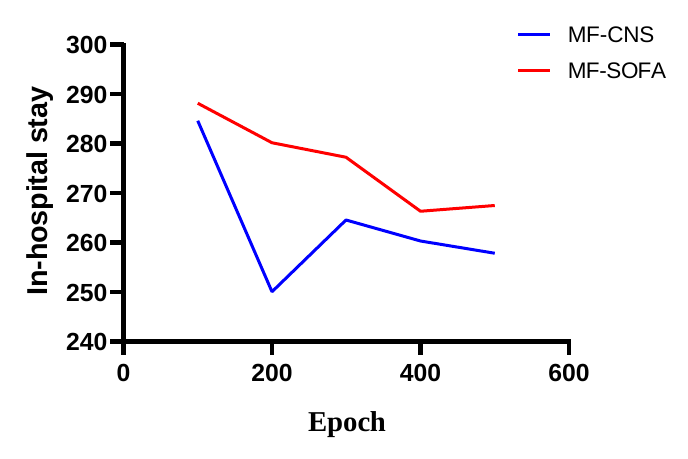}{d}{fig:hospital_d}
	\caption{Average length of ICU stays with MF-SOFA \& MF-CNS. Subfigure a represents the average length of ICU stay of DQN; Subfigure b represents the average length of ICU stay of DDQN; Subfigure c represents the average length of ICU stay of BCQ; Subfigure d represents the average length of ICU stay of CQL}
	\label{fig:hospital}
\end{figure}

We computed the two groups’ mortality rate and length of ICU stays concerning physicians’ actual treatment and various AI strategies. As shown in Fig. \ref{fig:Mortality}, Fig. \ref{fig:hospital}, and Table \ref{tab:mortality_icu_compact}, all models demonstrated improvements in reducing both mortality and length of stay, with the MF-CNS-based models consistently outperforming those based on MF-SOFA. Specifically, all results in this table were also subjected to the $T$-test with a $p$-value <0.05, indicating them statistically significant.

\begin{table}[htbp]
\caption{Mortality rates and average length of ICU stays of various strategies}
\centering
\label{tab:mortality_icu_compact}
\small
\resizebox{\textwidth}{!}{
\begin{tabular}{@{}lccccccccc@{}}
\toprule
\multirow{4}{*}{\centering Metric} & \multicolumn{9}{c}{Strategy} \\
\cmidrule(l){2-10}
 & \multirow{2}{*}{\centering PAT} & \multicolumn{2}{c}{DQN-MF-} & \multicolumn{2}{c}{DDQN-MF-} & \multicolumn{2}{c}{BCQ-MF-} & \multicolumn{2}{c}{CQL-MF-} \\
\cmidrule(lr){3-4} \cmidrule(lr){5-6} \cmidrule(lr){7-8} \cmidrule(lr){9-10} 
& & SOFA &  CNS  & SOFA &  CNS & SOFA &  CNS  & SOFA &  CNS \\
\midrule
\multirow{2}{*}{\centering MoR} & \multirow{2}{*}{\centering 13.24} & 9.76 &
 9.35 & 9.16 & 8.50 & 7.52 & 7.23 & 7.12 & \textbf{5.00}\\
 & & $\uparrow$29.38\% & $\uparrow$30.82\% &$\uparrow$26.28\% & $\uparrow$35.77\% & $\uparrow$43.20\% & $\uparrow$45.39\% & $\uparrow$46.22\% & \textbf{$\uparrow$62.24\% } \\
\multirow{2}{*}{\centering AiHS} & \multirow{2}{*}{\centering 308.48} & 
 279.09 & 275.59 & 275.83 & 271.82 & 274.6 & 254.78 & 266.35 & \textbf{250.07} \\
  & & $\uparrow$9.53\% & $\uparrow$10.66\% & $\uparrow$10.58\% & $\uparrow$11.88\% &$\uparrow$10.98\% & $\uparrow$17.41\% & $\uparrow$13.66\% & \textbf{$\uparrow$18.93\%} \\
\bottomrule
\end{tabular}
}
\parbox{0.98\columnwidth}{
\vspace{4pt}
\footnotesize\noindent\justifying
PAT, MoR, and AiHS denote the physicians' actual treatment, mortality rate, and average length of ICU stay, respectively. 
$\uparrow$ indicates the improvement compared to the actual value in PAT, calculated as: $(\text{MoR}_{\text{PAT}} - \text{MoR}_{\text{AI}})/\text{MoR}_{\text{PAT}}$, or $(\text{AiHS}_{\text{PAT}} - \text{AiHS}_{\text{AI}})/\text{AiHS}_{\text{PAT}}$, where AI denotes DQN-MF-SOFA, DQN-MF-CNS, DDQN-MF-SOFA, DDQN-MF-CNS, BCQ-MF-SOFA, BCQ-MF-CNS, CQL-MF-SOFA, CQL-MF-CNS.
}
\end{table}

Therefore, the comprehensive score CNS is significant. Among these AI strategies, CQL-MF-CNS has the best
performance, which improves the mortality rate from 13.24\% to 5.00\% and decreases the average length of ICU stay from 308.48 to 250.07 hours, the performances improved 62.24\% and 18.93\%. The results are reported in the last two rows of Table \ref{tab:mf_sofa_cns_performances}. 

\begin{mytable}[htbp]
\caption{Performances of MF-SOFA \& MF-CNS models with $\tau = 0.65$}
\label{tab:mf_sofa_cns_performances}
\centering
\resizebox{\textwidth}{!}{
\begin{tabular}{@{}lccccccccc@{}}
\toprule
\multirow{2}{*}{\centering Metric} & \multirow{2}{*}{\centering PAT} & \multicolumn{2}{c}{DQN-MF-} & \multicolumn{2}{c}{DDQN-MF-} & \multicolumn{2}{c}{BCQ-MF-} & \multicolumn{2}{c}{CQL-MF-} \\
\cmidrule(lr){3-4} \cmidrule(lr){5-6} \cmidrule(lr){7-8} \cmidrule(lr){9-10}
 & & SOFA & CNS  &  SOFA  & CNS  & SOFA  & CNS  &  SOFA  & CNS  \\
\midrule
OG & $\times$ & 0.781 & 0.600 & 0.595 & 0.581 & 0.647 & 0.632 & 0.628 & 0.573 \\
OB & $\times$ & 0.627 & 0.401 & 0.559 & 0.388 & 0.479 & 0.471 & 0.458 & 0.464 \\
WG & $\times$ & 0.361 & 0.4 & 0.429 & 0.297 & 0.393 & 0.299 & 0.213 & 0.250 \\
WB & $\times$ & 0.625 & 0.597 & 0.547 & 0.652 & 0.649 & 0.631 & 0.567 & 0.665 \\
$\sigma$ & $\times$ & 1.403 & 1.182 & 1.139 & 1.141 & 1.328 & 1.382 & 1.699 & 1.610 \\
$\mu$ & $\times$ & 0.409 & 0.421 & 0.153 & 0.112 & 0.092 & 0.114 & 0.196 & 0.122 \\
\rowcolor{yellow!35}
BE & $\times$ & 500 & 300 & 400 & 200 & 400 & 400 & 400 & 200 \\
\rowcolor{green!35}
MoR & 13.24 & 9.76 & 9.35 & 9.16 & 8.50 & 7.52 & 7.23 & 7.12 & \textbf{5.00} \\
\rowcolor{green!35}
AiHS & 308.48 & 279.09 & 275.59 & 275.83 & 271.82 & 274.6 & 254.78 & 266.35 & \textbf{250.07} \\
\bottomrule
\end{tabular}
}
\parbox{0.98\columnwidth}{
\vspace{4pt}
\footnotesize\noindent\justifying
PAT, BE, MoR, and AiHS denote the physicians' actual treatment, best epoch, mortality rate, and average length of ICU stay, respectively.
}
\end{mytable}

\begin{figure}[!htbp]
    \centering
{ \footnotesize
\begin{tabular}{cl cl cl cl}
{ \rule[0.5ex]{0.5cm}{0.8pt}} & DQN-MF-CNS & { \rule[0.5ex]{0.5cm}{0.8pt}} & DDQN-MF-CNS & { \rule[0.5ex]{0.5cm}{0.8pt}} & BCQ-MF-CNS & { \rule[0.5ex]{0.5cm}{0.8pt}} & CQL-MF-CNS \\
{ \rule[0.5ex]{0.2cm}{0.8pt} \rule[0.5ex]{0.2cm}{0.8pt}} & DQN-MF-SOFA & { \rule[0.5ex]{0.2cm}{0.8pt} \rule[0.5ex]{0.2cm}{0.8pt}} & DDQN-MF-SOFA & { \rule[0.5ex]{0.2cm}{0.8pt} \rule[0.5ex]{0.2cm}{0.8pt}} & BCQ-MF-SOFA & { \rule[0.5ex]{0.2cm}{0.8pt} \rule[0.5ex]{0.2cm}{0.8pt}} & CQL-MF-SOFA  
\end{tabular}
}

    \SubFigTL{0.50\textwidth}{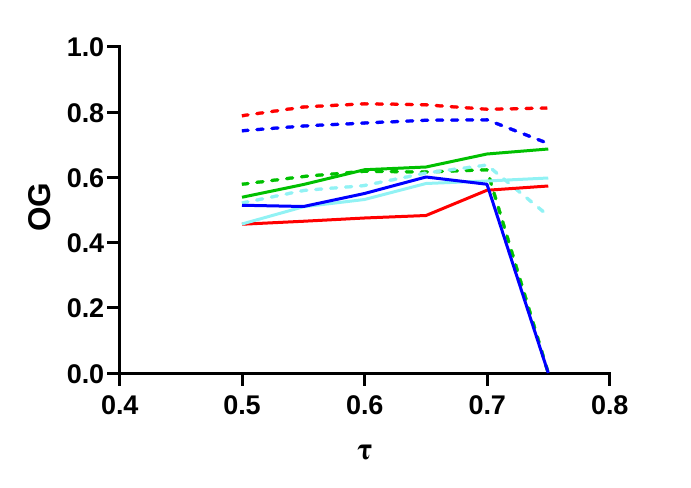}{a}{fig:epoch_a}
    \hfill
    \SubFigTL{0.48\textwidth}{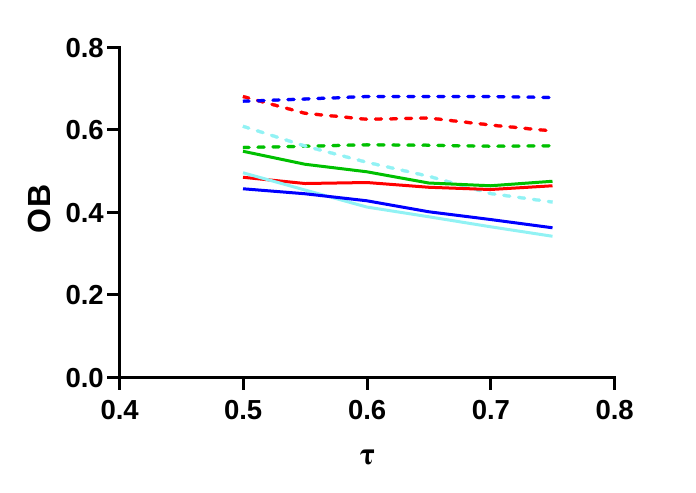}{b}{fig:epoch_b}
    \SubFigTL{0.48\textwidth}{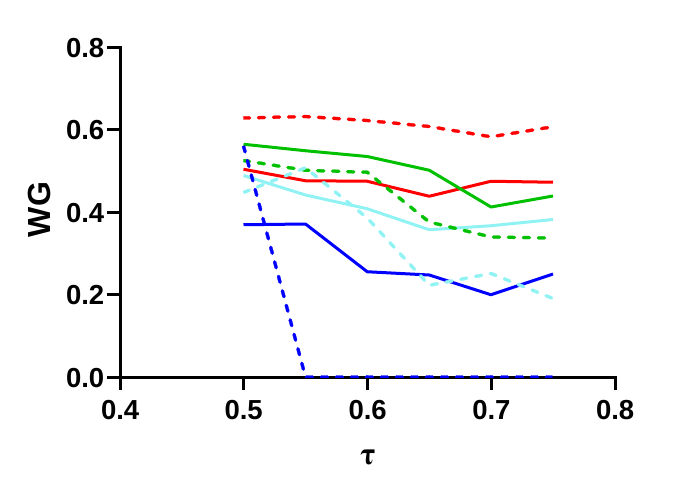}{c}{fig:epoch_c}
    \hfill
    \SubFigTL{0.48\textwidth}{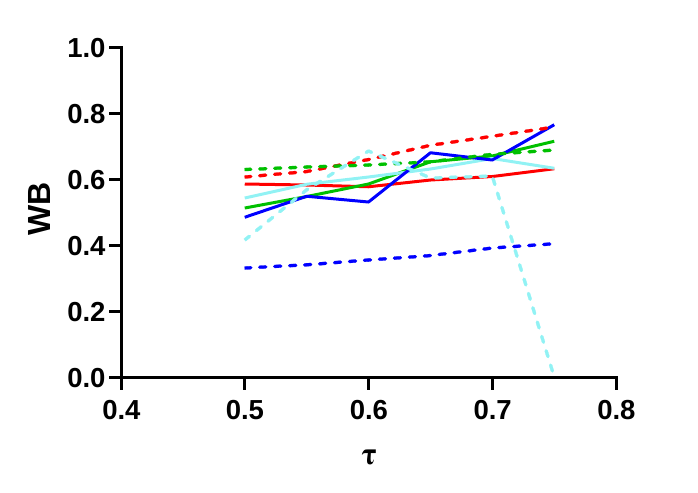}{d}{fig:epoch_d}
    \vspace{4mm}
    \SubFigTL{0.48\textwidth}{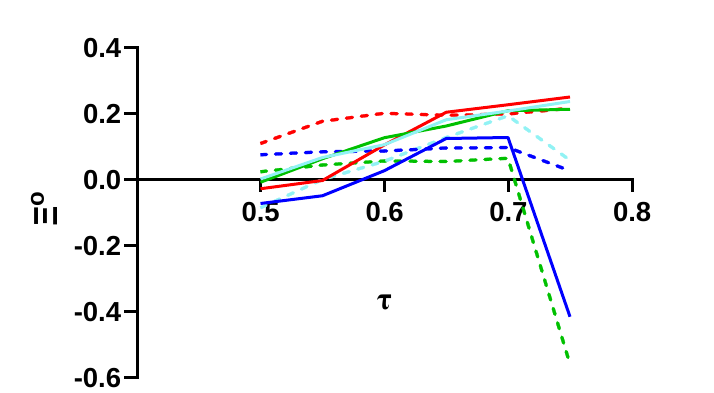}{e}{fig:epoch_e}
    \hfill
    \SubFigTL{0.48\textwidth}{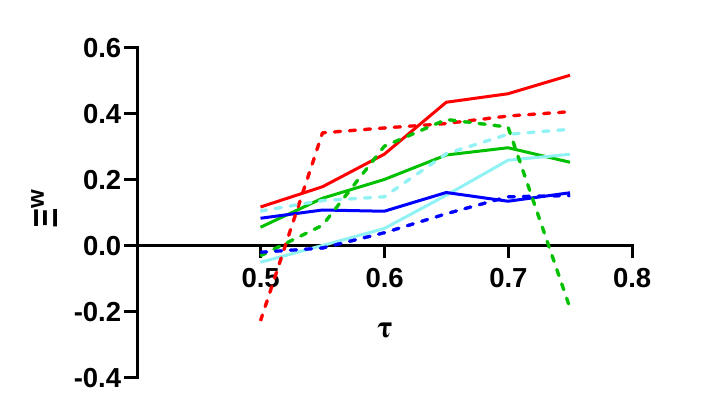}{f}{fig:epoch_f}
    \SubFigTL{0.48\textwidth}{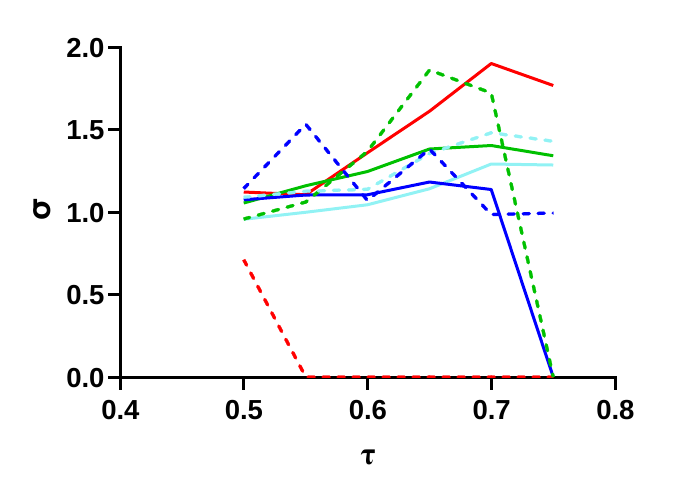}{g}{fig:epoch_g}
    \hfill
    \SubFigTL{0.48\textwidth}{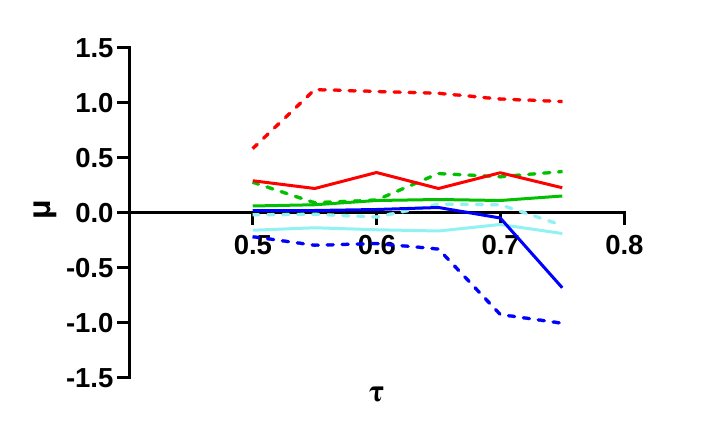}{h}{fig:epoch_h}
    \caption{TECM* of models via action similarity rate. Subfigure a-h represent the results for OG, OB, WG, WB, O-Gap $\Xi^o$, W-Gap $\Xi^w$, comprehensive confidence $\sigma$, and comprehensive bias $\mu$ of eight models}
    \label{fig:epoch_CQL}
\end{figure}

A second assessment method is the TECM*; we evaluate the performances of those eight models for the hyperparameter $\tau \in [0.5, 0.75]$, based on the external testing datasets with the same underlying episodes. In the evaluation, we chose the action similarity rate in \cite{liu2024value} as the basic metric. The result is presented in Fig. \ref{fig:epoch_CQL}, which indicates that the optimal AI strategies improve the physicians’ actual treatment, and each MF-CNS-version model improves the MF-SOFA-version model. Moreover, among those models, CQL-MF-CNS offers the best optimal AI strategies with the best fitting around $\tau = 0.65$.

\subsection{Ablation experiment}
To thoroughly investigate the impact of different state representations on the performance of the CQL-MF-CNS model, we designed and conducted a systematic ablation study. The CQL-MF-CNS algorithm was employed as the baseline model, with its architecture and hyperparameters held constant across all experimental settings. The model state comprises 19 vital signs and 4 RRT features: R1, R2, R3, R4. While maintaining constant vital signs dimensionality, we progressively evaluated 0- to 4-dimensional RRT feature combinations for mortality and length-of-stay prediction.

\begin{figure}[htp]
    \centering
    \begin{subfigure}[t]{0.48\textwidth}
        \centering
        \begin{tikzpicture}
            \node[anchor=south west,inner sep=0] (image) at (0,0) {
                \includegraphics[width=\textwidth]{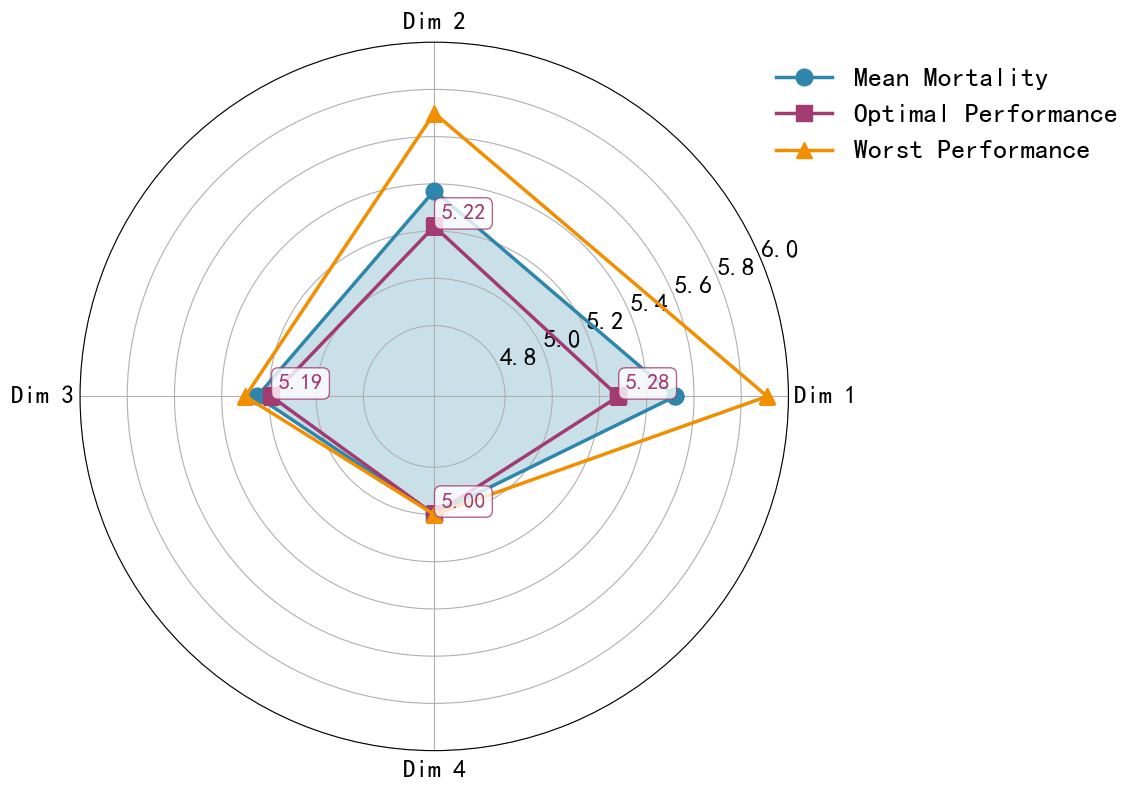}
            };
            \node[anchor=north west, xshift=5pt, yshift=-5pt, 
                  font=\small, text=black] at (image.north west) {a};
        \end{tikzpicture}
        \label{fig:left}
    \end{subfigure}
    \hfill
    \begin{subfigure}[t]{0.48\textwidth}
        \centering
        \begin{tikzpicture}
            \node[anchor=south west,inner sep=0] (image) at (0,0) {
                \includegraphics[width=\textwidth]{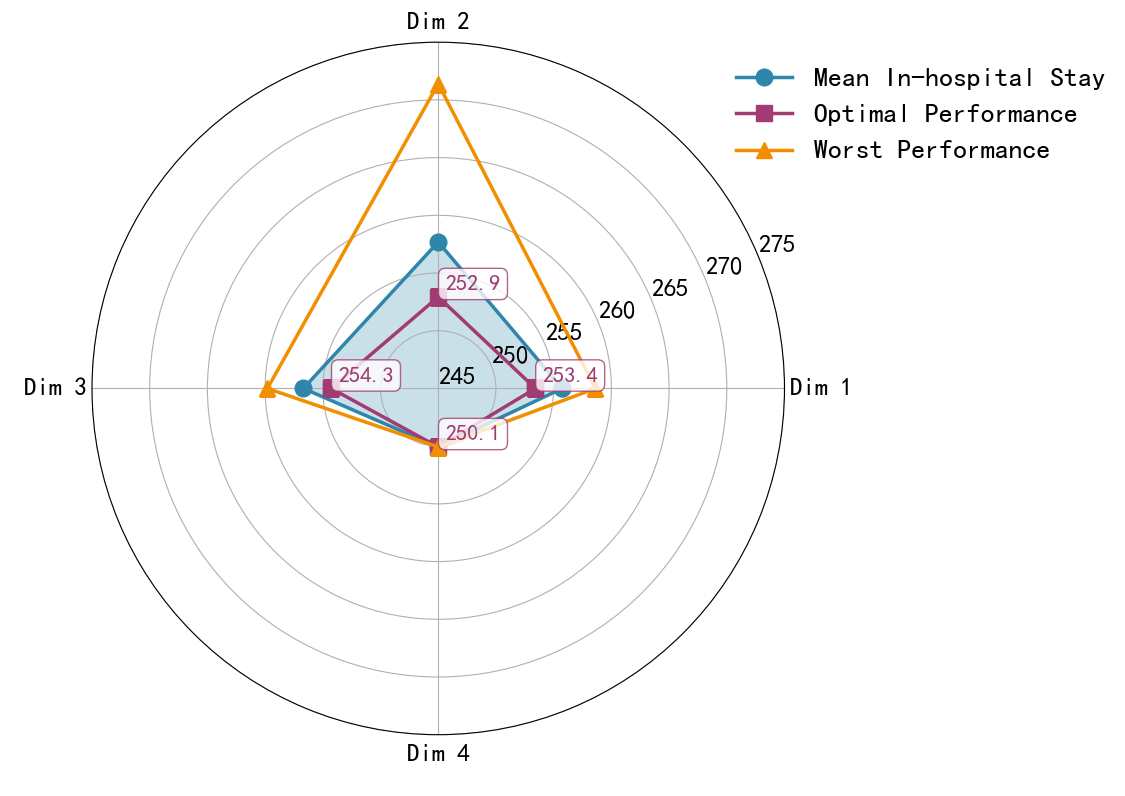}
            };
            \node[anchor=north west, xshift=5pt, yshift=-5pt, 
                  font=\small, text=black] at (image.north west) {b};
        \end{tikzpicture}
        \label{fig:right}
    \end{subfigure}
    
    \caption{Mortality and hospital stay of the CQL-MF-CNS model with varied RRT information dimensions (lower values indicate better performance)}
    \label{fig:Ablation_CQL}
\end{figure}

The results are presented in Table \ref{tab:rrt_combinations} and Fig. \ref{fig:Ablation_CQL}.
\begin{itemize}
    \item Mortality prediction: Increasing RRT dimensions from 0D to 4D reduced average mortality from 5.52\% to 5.00\%, with optimal/worst performance converging from 5.28\%/5.91\% to 5.19\%/5.30\%.
    \item Length-of-stay prediction: Average stay decreased from 253.79 hours (0D) to 250.07 hours (4D).
\end{itemize}

The dimensional expansion of RRT features significantly enhances model performance, demonstrating the
critical role of state space completeness in clinical decision-making.

\begin{table}[!htbp]
\centering
\caption{Performance of CQL model with different dimensional combinations of RRT features}
\label{tab:rrt_combinations}
\resizebox{\textwidth}{!}{
\begin{tabular}{>{\centering\arraybackslash}p{60pt}
>{\centering\arraybackslash}p{40pt}
>{\centering\arraybackslash}p{40pt}
>{\centering\arraybackslash}p{40pt}
>{\centering\arraybackslash}p{40pt}
>{\centering\arraybackslash}p{70pt}
>{\centering\arraybackslash}p{78pt}}
\toprule
\multirow{2}{*}{Dimension} & \multicolumn{4}{c}{Contains} & \multirow{2}{*}{Mortality} & \multirow{2}{*}{In-hospital stay} \\
\cmidrule(lr){2-5}
 & R1 & R2 & R3 & R4 & &  \\
\midrule
0 & 0 & 0 & 0 & 0 & 5.31 & 253.79 \\
\hline
\multirow{4}{*}{1} & 1 & 0 & 0 & 0 & 5.57 & 253.98 \\
 & 0 & 1 & 0 & 0 & 5.91 & 258.62 \\
 & 0 & 0 & 1 & 0 & 5.32 & \textbf{253.42} \\
 & 0 & 0 & 0 & 1 & \textbf{5.28} & 256.95 \\
\hline
\multirow{6}{*}{2} & 1 & 1 & 0 & 0 & 5.70 & 259.19 \\
 & 1 & 0 & 1 & 0 & 5.25 & 271.40 \\
 & 1 & 0 & 0 & 1 & 5.31 & 253.87 \\
 & 0 & 1 & 1 & 0 & 5.28 & \textbf{252.93} \\
 & 0 & 1 & 0 & 1 & \textbf{5.22} & 253.83 \\
 & 0 & 0 & 1 & 1 & 5.47 & 254.90 \\
\hline
\multirow{4}{*}{3} & 1 & 1 & 1 & 0 & \textbf{5.19} & 255.18 \\
 & 1 & 0 & 1 & 1 & 5.23 & 259.80 \\
 & 1 & 1 & 0 & 1 & 5.28 & 257.49 \\
 & 0 & 1 & 1 & 1 & 5.30 & \textbf{254.27} \\
\hline
4 & 1 & 1 & 1 & 1 & \textbf{5.00} & \textbf{250.07} \\
\bottomrule 
\end{tabular}
}
\parbox{0.98\columnwidth}{
\vspace{4pt}
\footnotesize\noindent\justifying
\footnotesize \justifying \noindent
This table systematically presents the performance of the CQL model on key metrics, including mortality prediction and length of stay prediction, when using different dimensional combinations of RRT features, given a fixed input of 19-dimensional vital signs. RRT features are encoded using a quadruple notation, where, for instance, ``1000'' denotes the use of the R1 feature only, and ``1100'' denotes the combination of R1 and R2 features.
}
\end{table}

\section{Discussion}

This study proposes a reinforcement learning approach based on a novel CNA to optimize RRT. A primary challenge in RRT practice is accurate patient status assessment. Unlike traditional scores using limited parameters, CNA integrates all available vital signs, providing a more comprehensive evaluation. CNA not only offers an intuitive clinical aid via its metric vector \((d_r, d_m, d_e)\), but its comprehensiveness also makes it suitable as a reward function for RL.Although CNA/CNS improves interpretability at the health-status assessment and reward-design levels, the learned reinforcement learning strategy itself is not fully interpretable. In particular, the model does not provide a complete clinical causal explanation for every recommendation of CRRT or IRRT. Therefore, the learned strategy should be regarded as a decision-support signal rather than an autonomous prescription rule.

We addressed the complexity of RRT strategy optimization by formulating the problem as a Markov Decision Process with a continuous state space.  In this formulation, each state incorporates demographic information, vital signs, and RRT-related treatment records, while the action space represents different RRT treatment choices. The reward function is defined using continuous CNS values to reflect changes in patient health status between adjacent time steps. Since the MDP framework models sequential decision-making under uncertainty, it is suitable for representing the dynamic evolution of patient conditions. Confronting the challenge of high missingness in real-world medical data, we further employed matrix factorization to reconstruct missing values, and the reconstruction results supported the effectiveness of this preprocessing step.

For evaluation, where real-world patient trials are ethically constrained, we utilized data-driven metrics including SCwHT \cite{wang2020sofa} and our proposed TECM. Applying the \(\eta\)-TECM termination criterion (\(\eta = 50\)), we identified the hyperparameter \(\tau = 0.65\) as optimally distinguishing ``good'' from ``bad'' actual physician treatments, thus using it as the boundary for AI strategy assessment. Ultimately, results in Tables \ref{tab:performances_compact} and \ref{tab:mf_sofa_cns_performances} demonstrate significant improvements achieved by the CNS-based AI strategies over actual treatments.

It should be noted that the present study was conducted on static retrospective cohorts from MIMIC-IV and eICU. Therefore, although the proposed MDP-based framework can model temporal patient trajectories and uncertain treatment effects, real-time non-stationary shifts and concept drift in clinical environments were not directly evaluated in this work. In practice, patient populations, clinical protocols, treatment resources, and disease progression patterns may change over time, which may affect the stability of a learned strategy. The adopted Q-type offline reinforcement learning algorithms, including DQN, DDQN, BCQ, and CQL, are trained with experience replay, where state-action-reward-transition samples are stored and reused for strategy optimization. This mechanism provides a natural basis for extending the current framework to incorporate newly collected clinical data and periodically update or recalibrate the learned strategy. However, such online adaptation was not implemented in the current retrospective validation and will be investigated in future work through drift detection, continual learning, and prospective real-time evaluation.

Another important safety issue concerns the extrapolation risk of offline reinforcement learning. In practical clinical applications, an AI-recommended RRT action may deviate from the historical treatment distribution, particularly when the model encounters rare or previously unobserved state-action combinations. Such unsupported combinations may lead to unreliable value estimation and unsafe strategy recommendations. To mitigate this issue, this study included offline RL algorithms such as BCQ and CQL. BCQ constrains the learned strategy toward actions that are likely under the behavior strategy, thereby reducing extrapolation to poorly supported actions. CQL further adopts a conservative value-estimation strategy to penalize potentially overestimated values for out-of-distribution actions. These mechanisms are consistent with the motivation of offline RL for reducing strategy bias and limiting unsafe deviations from historical clinical practice.

Nevertheless, BCQ and CQL can only reduce, rather than completely eliminate, the risk associated with extreme or unsupported state-action combinations. Therefore, the proposed framework should be interpreted as an exploratory offline algorithmic validation rather than a clinically deployable autonomous decision-making system. Before any real-world clinical use, the learned strategy should be combined with explicit clinical safety constraints, uncertainty estimation, out-of-distribution detection, and clinician-in-the-loop review. In addition, prospective validation and carefully monitored clinical studies would be necessary to evaluate whether the model can provide safe and reliable recommendations under real clinical conditions.

Beyond its current application in health status assessment, the concept of CNA may provide a new perspective for developing advanced normalization strategies in neural networks. Existing normalization methods, such as batch normalization and layer normalization, mainly regulate feature distributions based on statistical properties, while largely ignoring the intrinsic relationships, relative importance, and contextual meanings among different variables. In contrast, CNA introduces a comprehensive normalization paradigm by integrating multi-dimensional information to generate a more biologically and functionally meaningful representation space. Inspired by this principle, CNA could be further extended into a trainable neural network normalization layer that dynamically adjusts feature representations according to both data-driven characteristics and domain-specific knowledge. Such a CNA-based normalization approach may improve the robustness, interpretability, and generalizability of artificial intelligence models, particularly for complex biomedical datasets characterized by high dimensionality, heterogeneity, and individual variability. Therefore, future studies exploring the integration of CNA principles with deep learning architectures may establish a novel normalization framework that bridges conventional data preprocessing and adaptive representation learning.

\section{Conclusion}
The findings of this study indicate that the combination of the novel CNA with Reinforcement Learning methods offers a promising direction for optimizing RRT. By addressing key challenges related to state definition, reward function design, handling missing data, and evaluating AI strategies, our developed Q-learning-based approach demonstrates significant performance improvements over actual physician strategies in data-driven assessments.  In retrospective assessments using real-world clinical databases, the proposed approach showed performance improvements over actual physician strategies. However, these findings should be interpreted as pre-clinical algorithmic validation rather than evidence for immediate clinical implementation. Prospective real-world validation and rigorously designed clinical trials are required before the framework can be applied in clinical practice. Future work should further integrate expert clinical knowledge, strengthen safety and interpretability analyses, and evaluate the framework under appropriate ethical approval, safety monitoring, and clinician supervision. Beyond its current application, CNA may open new avenues for designing next-generation neural network normalization approaches, bridging comprehensive data assessment with adaptive artificial intelligence modeling.

\bibliographystyle{plain}

\bibliography{cas-refs_r}

@ARTICLE{xu2024narrative,
  title={A narrative review on the application of artificial intelligence in renal ultrasound},
  author={Xu, Tong and Zhang, Xian-Ya and Yang, Na and Jiang, Fan and Chen, Gong-Quan and Pan, Xiao-Fang and Peng, Yue-Xiang and Cui, Xin-Wu},
  journal={Frontiers in Oncology},
  volume={13},
  pages={1252630},
  year={2024}
}

@ARTICLE{kelly2021potential,
  title={The potential for artificial intelligence to predict clinical outcomes in patients who have acquired acute kidney injury during the perioperative period},
  author={Kelly, Barry J and Chevarria, Julio and O’Sullivan, Barry and Shorten, George},
  journal={Perioperative Medicine},
  volume={10},
  number={1},
  pages={49},
  year={2021},
  publisher={Springer}
}

@ARTICLE{hammouda2022can,
  title={Can Augmented Intelligence Assist in Delivering Continuous Renal Replacement Therapy? A Scoping Review: {FR-PO102}},
  author={Hammouda, Nada and Neyra, Javier A},
  journal={Journal of the American Society of Nephrology},
  volume={33},
  number={11S},
  pages={354},
  year={2022},
  publisher={LWW}
}

@ARTICLE{yoo2023predicting,
  title={Predicting outcomes of continuous renal replacement therapy using body composition monitoring: a deep-learning approach},
  author={Yoo, Kyung Don and Noh, Junhyug and Bae, Wonho and An, Jung Nam and Oh, Hyung Jung and Rhee, Harin and Seong, Eun Young and Baek, Seon Ha and Ahn, Shin Young and Cho, Jang-Hee and others},
  journal={Scientific Reports},
  volume={13},
  number={1},
  pages={4605},
  year={2023},
  publisher={Nature Publishing Group UK London}
}

@ARTICLE{leung2024deep,
  title={Deep learning algorithms for predicting renal replacement therapy initiation in {CKD} patients: a retrospective cohort study},
  author={Leung, Ka-Chun and Ng, Wincy Wing-Sze and Siu, Yui-Pong and Hau, Anthony Kai-Ching and Lee, Hoi-Kan},
  journal={BMC Nephrology},
  volume={25},
  number={1},
  pages={95},
  year={2024},
  publisher={Springer}
}

@ARTICLE{schoenfelder2017effects,
  title={Effects of continuous and intermittent renal replacement therapies among adult patients with acute kidney injury},
  author={Schoenfelder, Tonio and Chen, Xiaoyu and Ble{\ss}, Hans-Holger},
  journal={GMS Health Technology Assessment},
  volume={13},
  pages={Doc01},
  year={2017}
}

@ARTICLE{lins2009intermittent,
  title={Intermittent versus continuous renal replacement therapy for acute kidney injury patients admitted to the intensive care unit: results of a randomized clinical trial},
  author={Lins, Robert L and Elseviers, Monique M and Van der Niepen, Patricia and Hoste, Eric and Malbrain, Manu L and Damas, Pierre and Devriendt, Jacques and {SHARF investigators}},
  journal={Nephrology Dialysis Transplantation},
  volume={24},
  number={2},
  pages={512--518},
  year={2009},
  publisher={Oxford University Press}
}

@ARTICLE{wang2024clinical,
  title={Clinical knowledge-guided deep reinforcement learning for sepsis antibiotic dosing recommendations},
  author={Wang, Yuan and Liu, Anqi and Yang, Jucheng and Wang, Lin and Xiong, Ning and Cheng, Yisong and Wu, Qin},
  journal={Artificial Intelligence in Medicine},
  volume={150},
  pages={102811},
  year={2024},
  publisher={Elsevier}
}

@ARTICLE{qayyum2020secure,
  title={Secure and robust machine learning for healthcare: A survey},
  author={Qayyum, Adnan and Qadir, Junaid and Bilal, Muhammad and Al-Fuqaha, Ala},
  journal={IEEE Reviews in Biomedical Engineering},
  volume={14},
  pages={156--180},
  year={2020},
  publisher={IEEE}
}

@ARTICLE{lee2018machine,
  title={Machine learning: Overview of the recent progresses and implications for the process systems engineering field},
  author={Lee, Jay H and Shin, Joohyun and Realff, Matthew J},
  journal={Computers {\upshape\&} Chemical Engineering},
  volume={114},
  pages={111--121},
  year={2018},
  publisher={Elsevier}
}

@ARTICLE{ando2005framework,
  title={A framework for learning predictive structures from multiple tasks and unlabeled data.},
  author={Ando, Rie Kubota and Zhang, Tong and Bartlett, Peter},
  journal={Journal of Machine Learning Research},
  volume={6},
  number={11},
  year={2005}
}

@ARTICLE{khamis2014adaptive,
  title={Adaptive multi-objective reinforcement learning with hybrid exploration for traffic signal control based on cooperative multi-agent framework},
  author={Khamis, Mohamed A and Gomaa, Walid},
  journal={Engineering Applications of Artificial Intelligence},
  volume={29},
  pages={134--151},
  year={2014},
  publisher={Elsevier}
}

@ARTICLE{lewis2012reinforcement,
  title={Reinforcement learning and feedback control: Using natural decision methods to design optimal adaptive controllers},
  author={Lewis, Frank L and Vrabie, Draguna and Vamvoudakis, Kyriakos G},
  journal={IEEE Control Systems Magazine},
  volume={32},
  number={6},
  pages={76--105},
  year={2012},
  publisher={IEEE}
}

@ARTICLE{kuhnle2021designing,
  title={Designing an adaptive production control system using reinforcement learning},
  author={Kuhnle, Andreas and Kaiser, Jan-Philipp and Thei{\ss}, Felix and Stricker, Nicole and Lanza, Gisela},
  journal={Journal of Intelligent Manufacturing},
  volume={32},
  number={3},
  pages={855--876},
  year={2021},
  publisher={Springer}
}

@ARTICLE{liu2024value,
  title={Value function assessment to different {RL} algorithms for heparin treatment policy of patients with sepsis in {ICU}},
  author={Liu, Jiang and Xie, Yihao and Shu, Xin and Chen, Yuwen and Sun, Yizhu and Zhong, Kunhua and Liang, Hao and Li, Yujie and Yang, Chunyong and Han, Yan and others},
  journal={Artificial Intelligence in Medicine},
  volume={147},
  pages={102726},
  year={2024},
  publisher={Elsevier}
}

@ARTICLE{vincent1996sofa,
  title={The {SOFA} (Sepsis-related Organ Failure Assessment) score to describe organ dysfunction/failure: On behalf of the Working Group on Sepsis-Related Problems of the European Society of Intensive Care Medicine (see contributors to the project in the appendix)},
  author={Vincent, Jean-Louis and Moreno, Rui and Takala, Jukka and Willatts, Sheila and De Mendon{\c{c}}a, Arnaldo and Bruining, Hajo and Reinhart, C K and Suter, Peter M and Thijs, Lambertius G},
  journal={Intensive Care Medicine},
  volume={22},
  number={7},
  pages={707--710},
  year={1996},
  publisher={Springer-Verlag Berlin/Heidelberg}
}

@ARTICLE{lambden2019sofa,
  title={The {SOFA} score—development, utility and challenges of accurate assessment in clinical trials},
  author={Lambden, Simon and Laterre, Pierre Francois and Levy, Mitchell M and Francois, Bruno},
  journal={Critical Care},
  volume={23},
  number={1},
  pages={374},
  year={2019},
  publisher={Springer}
}

@ARTICLE{knaus1985apache,
  title={{APACHE} {II}: a severity of disease classification system},
  author={Knaus, William A and Draper, Elizabeth A and Wagner, Douglas P and Zimmerman, Jack E},
  journal={Critical Care Medicine},
  volume={13},
  number={10},
  pages={818--829},
  year={1985},
  publisher={LWW}
}

@ARTICLE{awdishu2025kdigo,
  title={{KDIGO} 2024 clinical practice guideline on evaluation and management of chronic kidney disease: A primer on what pharmacists need to know},
  author={Awdishu, Linda and Maxson, Rebecca and Gratt, Chelsea and Rubenzik, Tamara and Battistella, Marisa},
  journal={American Journal of Health-System Pharmacy},
  volume={82},
  number={12},
  pages={660--671},
  year={2025},
  publisher={Oxford University Press US}
}

@ARTICLE{kotani2019modification,
  title={Modification of sequential organ failure assessment score using acute kidney injury classification},
  author={Kotani, Yuki and Fujii, Tomoko and Uchino, Shigehiko and Doi, Kent and {JAKID Study Group}},
  journal={Journal of Critical Care},
  volume={51},
  pages={198--203},
  year={2019},
  publisher={Elsevier}
}

@ARTICLE{le1993new,
  title={A new simplified acute physiology score ({SAPS} {II}) based on a {European}/{North American} multicenter study},
  author={Le Gall, Jean-Roger and Lemeshow, Stanley and Saulnier, Fabienne},
  journal={JAMA},
  volume={270},
  number={24},
  pages={2957--2963},
  year={1993},
  publisher={American Medical Association}
}

@ARTICLE{bae2008continuous,
  title={Continuous renal replacement therapy for the treatment of acute kidney injury},
  author={Bae, Woo Kyun and Lim, Dae Hun and Jeong, Ji Min and Jung, Hae Young and Kim, Seong Ku and Park, Jeong Woo and Bae, Eun Hui and Ma, Seong Kwon and Kim, Soo Wan and Kim, Nam Ho and others},
  journal={The Korean Journal of Internal Medicine},
  volume={23},
  number={2},
  pages={58},
  year={2008}
}

@ARTICLE{ricci2008rifle,
  title={The {RIFLE} criteria and mortality in acute kidney injury: a systematic review},
  author={Ricci, Zaccaria and Cruz, Dinna and Ronco, Claudio},
  journal={Kidney International},
  volume={73},
  number={5},
  pages={538--546},
  year={2008},
  publisher={Elsevier}
}

@ARTICLE{mehta2007acute,
  title={{Acute Kidney Injury Network}: report of an initiative to improve outcomes in acute kidney injury},
  author={Mehta, Ravindra L and Kellum, John A and Shah, Sudhir V and Molitoris, Bruce A and Ronco, Claudio and Warnock, David G and Levin, Adeera and {Acute Kidney Injury Network}},
  journal={Critical Care},
  volume={11},
  number={2},
  pages={R31},
  year={2007},
  publisher={Springer}
}

@ARTICLE{BEOLET2024102920,
title = {End-to-end offline reinforcement learning for glycemia control},
journal = {Artificial Intelligence in Medicine},
volume = {154},
pages = {102920},
year = {2024},
issn = {0933-3657},
doi = {https://doi.org/10.1016/j.artmed.2024.102920},
url = {https://www.sciencedirect.com/science/article/pii/S0933365724001623},
author = {Beolet, Tristan and Adenis, Alice and Huneker, Erik and Louis, Maxime}
}

@INPROCEEDINGS{fujimoto2019off,
  title={Off-policy deep reinforcement learning without exploration},
  author={Fujimoto, Scott and Meger, David and Precup, Doina},
  booktitle={International Conference on Machine Learning},
  pages={2052--2062},
  year={2019},
  organization={PMLR}
}

@ARTICLE{kumar2020conservative,
  title={Conservative {Q}-learning for offline reinforcement learning},
  author={Kumar, Aviral and Zhou, Aurick and Tucker, George and Levine, Sergey},
  journal={Advances in Neural Information Processing Systems},
  volume={33},
  pages={1179--1191},
  year={2020}
}

@ARTICLE{mnih2013playing,
  title={Playing {Atari} with deep reinforcement learning},
  author={Mnih, Volodymyr and Kavukcuoglu, Koray and Silver, David and Graves, Alex and Antonoglou, Ioannis and Wierstra, Daan and Riedmiller, Martin},
  journal={arXiv preprint arXiv:1312.5602},
  year={2013}
}

@ARTICLE{hasselt2010double,
  title={Double {Q}-learning},
  author={Hasselt, Hado},
  journal={Advances in Neural Information Processing Systems},
  volume={23},
  year={2010}
}

@INPROCEEDINGS{kuang2018performance,
  title={Performance dynamics and termination errors in reinforcement learning--a unifying perspective},
  author={Kuang, Nikki Lijing and Leung, Clement HC},
  booktitle={2018 IEEE First International Conference on Artificial Intelligence and Knowledge Engineering (AIKE)},
  pages={129--133},
  year={2018},
  organization={IEEE}
}

@ARTICLE{nguyen2019low,
  title={Low-rank matrix completion: A contemporary survey},
  author={Nguyen, Luong Trung and Kim, Junhan and Shim, Byonghyo},
  journal={IEEE Access},
  volume={7},
  pages={94215--94237},
  year={2019},
  publisher={IEEE}
}

@ARTICLE{wu2019deep,
  title={A deep latent factor model for high-dimensional and sparse matrices in recommender systems},
  author={Wu, Di and Luo, Xin and Shang, Mingsheng and He, Yi and Wang, Guoyin and Zhou, MengChu},
  journal={IEEE Transactions on Systems, Man, and Cybernetics: Systems},
  volume={51},
  number={7},
  pages={4285--4296},
  year={2019},
  publisher={IEEE}
}

@ARTICLE{song2019tensor,
  title={Tensor completion algorithms in big data analytics},
  author={Song, Qingquan and Ge, Hancheng and Caverlee, James and Hu, Xia},
  journal={ACM Transactions on Knowledge Discovery from Data (TKDD)},
  volume={13},
  number={1},
  pages={1--48},
  year={2019},
  publisher={ACM New York, NY, USA}
}

@ARTICLE{liu2025tecm,
  title={{TECM*}: A Data-Driven Assessment to Reinforcement Learning Methods and Application to Heparin Treatment Strategy for Surgical Sepsis},
  author={Liu, Jiang and Li, Yujie and Zhou, Chan and Xie, Yihao and Sun, Qilong and Shu, Xin and Li, Peiwei and Yang, Chunyong and Zhu, Yiziting and Zhu, Jiaqi and others},
  journal={arXiv preprint arXiv:2512.10973},
  year={2025}
}

@ARTICLE{ostermann2016acute,
  title={Acute kidney injury 2016: diagnosis and diagnostic workup},
  author={Ostermann, Marlies and Joannidis, Michael},
  journal={Critical Care},
  volume={20},
  number={1},
  pages={299},
  year={2016},
  publisher={Springer}
}

@ARTICLE{hou2023effects,
  title={Effects of continuous renal replacement therapy on {APACHE}-{II} score, creatinine, and urea nitrogen Levels in patients with acute kidney injury},
  author={Hou, Huihui and Li, Lingzhi},
  journal={Pakistan Journal of Medical Sciences},
  volume={39},
  number={1},
  pages={50},
  year={2023}
}

@ARTICLE{johnson2023mimic,
  title={{MIMIC}-{IV}, a freely accessible electronic health record dataset},
  author={Johnson, Alistair EW and Bulgarelli, Lucas and Shen, Lu and Gayles, Alvin and Shammout, Ayad and Horng, Steven and Pollard, Tom J and Hao, Sicheng and Moody, Benjamin and Gow, Brian and others},
  journal={Scientific Data},
  volume={10},
  number={1},
  pages={1},
  year={2023},
  publisher={Nature Publishing Group UK London}
}

@ARTICLE{li2020optimizing,
  title={Optimizing medical treatment for sepsis in intensive care: from reinforcement learning to pre-trial evaluation},
  author={Li, Luchen and Albert-Smet, Ignacio and Faisal, Aldo A},
  journal={arXiv preprint arXiv:2003.06474},
  year={2020}
}

@ARTICLE{raghu2017deep,
  title={Deep reinforcement learning for sepsis treatment},
  author={Raghu, Aniruddh and Komorowski, Matthieu and Ahmed, Imran and Celi, Leo and Szolovits, Peter and Ghassemi, Marzyeh},
  journal={arXiv preprint arXiv:1711.09602},
  year={2017}
}

@INPROCEEDINGS{werbos1999stable,
  title={Stable adaptive control using new critic designs},
  author={Werbos, Paul J},
  booktitle={Ninth workshop on virtual intelligence/dynamic neural networks},
  volume={3728},
  pages={510--579},
  year={1999},
  organization={SPIE}
}

@ARTICLE{oh2021optimal,
  title={Optimal treatment recommendations for diabetes patients using the {Markov} decision process along with the {South} {Korean} electronic health records},
  author={Oh, Sang-Ho and Lee, Su Jin and Noh, Juhwan and Mo, Jeonghoon},
  journal={Scientific Reports},
  volume={11},
  number={1},
  pages={6920},
  year={2021},
  publisher={Nature Publishing Group UK London}
}

@ARTICLE{wang2020sofa,
  title={{SOFA} score is superior to {APACHE}-{II} score in predicting the prognosis of critically ill patients with acute kidney injury undergoing continuous renal replacement therapy},
  author={Wang, Hai and Kang, Xiao and Shi, Yu and Bai, Zheng-hai and Lv, Jun-hua and Sun, Jiang-li and Pei, Hong-hong},
  journal={Renal Failure},
  volume={42},
  number={1},
  pages={638--645},
  year={2020},
  publisher={Taylor \& Francis}
}

@ARTICLE{kapral2025optimal,
  title={Optimal timing for renal replacement therapy in critically ill patients using reinforcement learning algorithms},
  author={Kapral, Lorenz and Azarbeik, Mohammad Mahdi and Weiss, Richard and Bologheanu, Razvan and Heitzinger, Clemens and Kimberger, Oliver},
  journal={Journal of Critical Care},
  volume={86},
  pages={154964},
  year={2025},
  publisher={Elsevier}
}

@ARTICLE{peine2021development,
  title={Development and validation of a reinforcement learning algorithm to dynamically optimize mechanical ventilation in critical care},
  author={Peine, Arne and Hallawa, Ahmed and Bickenbach, Johannes and Dartmann, Guido and Fazlic, Lejla Begic and Schmeink, Anke and Ascheid, Gerd and Thiemermann, Christoph and Schuppert, Andreas and Kindle, Ryan and others},
  journal={npj Digital Medicine},
  volume={4},
  number={1},
  pages={32},
  year={2021},
  publisher={Nature Publishing Group UK London}
}

@ARTICLE{zhang2025long,
  title={Long-term safety of “4-hour” hemoadsorption combined with hemodialysis in maintenance hemodialysis patients: a multicenter prospective cohort study},
  author={Zhang, Dongliang and Liu, Cuiping and Yang, Tao and Zhao, Jingxin and Wang, Xiaofei and Zhang, Liping and Li, Yuanyuan and Shen, Yangyang and Gao, Yanjun and Zhang, Hongjuan},
  journal={Blood Purification},
  volume={54},
  number={7},
  pages={413--423},
  year={2025},
  publisher={S. Karger AG}
}

@ARTICLE{saha2017effective,
  title={Effective sparse imputation of patient conditions in electronic medical records for emergency risk predictions},
  author={Saha, Budhaditya and Gupta, Sunil and Phung, Dinh and Venkatesh, Svetha},
  journal={Knowledge and Information Systems},
  volume={53},
  number={1},
  pages={179--206},
  year={2017},
  publisher={Springer}
}

@INPROCEEDINGS{wu2015post,
  title={Post-surgical complication prediction in the presence of low-rank missing data},
  author={Wu, Hang and Cheng, Chihwen and Han, Xiaoning and Huo, Yong and Ding, Wenhui and Wang, May D},
  booktitle={2015 37th Annual International Conference of the IEEE Engineering in Medicine and Biology Society (EMBC)},
  pages={6808--6811},
  year={2015},
  organization={IEEE}
}

\end{document}